%% file: acl_latex.tex
\pdfoutput=1

\documentclass[11pt]{article}

\usepackage[preprint]{acl}

\usepackage{times}
\usepackage{latexsym}

\usepackage[T1]{fontenc}

\usepackage[utf8]{inputenc}

\usepackage{microtype}

\usepackage{inconsolata}
\usepackage{tcolorbox}
\tcbuselibrary{breakable}
\usepackage{xcolor}

\usepackage{graphicx}
\input{premable}

\title{POINTS-Reader: Distillation-Free Adaptation of Vision-Language Models for Document Conversion}

\author{
  \hspace{-0.25cm}\textbf{Yuan Liu$^{1}$, Zhongyin Zhao$^{1}$, Le Tian$^{1}$, Haicheng Wang$^{1,2}$, Xubing Ye$^{1,3}$} \\
  \hspace{-0.25cm}\textbf{Yangxiu You$^{1}$, Zilin Yu$^{1}$, Chuhan Wu$^{1}$, Xiao Zhou$^{1}$, Yang Yu$^{1}$, Jie Zhou$^{1}$} \\
  \hspace{-0.25cm}$^{1}$Pattern Recognition Center, WeChat AI, Tencent Inc, China \\
  \hspace{-0.25cm}$^{2}$Shanghai Jiao Tong University, $^{3}$Tsinghua University \\
  \hspace{-0.25cm}\tt\small\{bensenliu\}@tencent.com \\
  }

\begin{document}
\maketitle
\input{text/0-abs}
\input{text/1-intro}

\input{text/3-methods}

\input{text/4-exps}
\input{text/2-related}
\input{text/5-con}

\bibliography{custom}

\clearpage

\appendix
\input{text/6-appendix}

\end{document}

%% file: premable.tex
\usepackage{wrapfig}
\usepackage{enumitem}
\usepackage{mmstyle}
\usepackage{color, colortbl}

\usepackage{algorithm}
\usepackage{algorithmic}
\usepackage{caption}
\usepackage{amsmath}
\usepackage{graphicx}
\usepackage{subcaption}
\usepackage{booktabs}
\usepackage{multirow} 
\usepackage{hyperref}
\usepackage{arydshln}

%% file: text/0-abs.tex
\begin{abstract}
    High-quality labeled data is essential for training accurate document conversion models, particularly in domains with complex formats such as tables, formulas, and multi-column text. However, manual annotation is both costly and time-consuming, while automatic labeling using existing models often lacks accuracy in handling such challenging scenarios. Consequently, training student models by distilling outputs from teacher models can significantly limit their performance in real-world applications. In this paper, we propose a fully automated, distillation-free framework comprising two stages for constructing high-quality document extraction datasets and models capable of handling diverse document formats and layouts. In the first stage, we introduce a method for generating large-scale, diverse synthetic data, which enables a model to extract key elements in a unified format with strong initial performance. In the second stage, we present a self-improvement approach that further adapts the model, initially trained on synthetic data, to real-world documents. Specifically, we first use the fine-tuned model to annotate real documents, then apply a suite of filtering strategies to verify annotation quality, and finally retrain the model on the verified dataset. By iteratively repeating this process, we progressively enhance both the model’s conversion capabilities and the quality of the generated data. We train a public POINTS-1.5 model to obtain \textit{POINTS-Reader}, which surpasses many existing public and proprietary models of comparable or larger size. Our model is available at \url{https://github.com/Tencent/POINTS-Reader}\footnote{Accepted by EMNLP2025 Main Conference}.
\end{abstract}

%% file: text/1-intro.tex
\section{Introduction}
The internet contains a vast and ever-expanding collection of publicly available documents, including textbooks, scientific articles, and technical reports. These resources encapsulate extensive world knowledge and are essential for pre-training large language models~\citep{yang2024qwen2, team2025kimivl, abdin2024phi}. However, accurately converting such documents into text—particularly for complex elements like tables and mathematical formulas—remains a significant challenge. Due to the scarcity of high-quality annotated datasets, most existing approaches~\citep{poznanski2025olmocr} address this issue by collecting large-scale document-image datasets using external models, and subsequently fine-tuning vision-language models~\citep{bai2025qwen2, chen2024expanding, hurst2024gpt} for end-to-end document conversion. This paradigm introduces two major issues. First, the reliance on external models hinders the research and development of next-generation models, and heavy dependence on distillation may obscure the true effectiveness of training vision-language models from scratch~\citep{cho2025perceptionlm}. Second, student models often fail to fully match the performance of teacher models and may also inherit their biases (\autoref{fig:fig1}(b)).

To overcome these limitations, it is necessary to construct datasets without relying on distillation from external models. Compared to labor-intensive manual annotation, generating large amounts of synthetic data appears to be a promising direction. However, due to the substantial differences between synthetic and real-world samples, further adaptation on real-world datasets is required. To this end, we propose a fully automated pipeline for constructing large-scale, high-quality document conversion datasets, consisting of two stages: the \textbf{U}niform format \textbf{W}arm-up \textbf{S}tage (\textbf{UWS}) and the \textbf{I}terative \textbf{S}elf-improvement \textbf{S}tage (\textbf{ISS}).

\input{figures/figure_1}

\paragraph{Uniform format Warm-up Stage} 

Documents contain a variety of elements, such as plain text, tables, and mathematical formulas, each requiring different output formats. For example, tables can be represented in Markdown, HTML, or LaTeX formats. This diversity increases the learning difficulty for models tasked with document understanding. To address this challenge, we first standardize the output formats for these elements, as detailed in Section~\ref{sec:method}. Guided by these unified output formats, we generate a large number of document texts using a large language model and render them into images. These image-text pairs are then used to fine-tune a general vision-language model, such as POINTS-1.5~\citep{liu2024points1}. This approach enables the model to accurately output plain text, tables, and formulas in a consistent format, thereby laying the foundation for more robust and generalizable document understanding capabilities.

\paragraph{Iterative Self-improvement Stage.} While the model trained in the previous stage can extract key elements from documents in a consistent format, its performance on real-world documents, especially those with complex layouts, remains suboptimal. To bridge this gap, we introduce an iterative self-improvement framework that enables the model to autonomously generate and refine large-scale, high-quality training data. Specifically, we apply the current model to generate textual annotations on large-scale real-world documents. However, these initial outputs often suffer from issues like missing main text, hallucinations\citep{nassar2025smoldocling}, incomplete table cells, and syntactic errors in mathematical formulas. To address these challenges, we design targeted filtering strategies to automatically validate the generated data. The refined dataset is then used to further retrain the model. By repeating this process over multiple iterations, both the model’s extraction accuracy and the quality of the generated data improve substantially. \autoref{fig:fig1}(c) presents a sample of the data generated at this stage.

Our contributions are summarized as follows:
\begin{enumerate}[label={\bf $\bullet$}, leftmargin=*, topsep=0.5ex, itemsep=-0.5ex, partopsep=0.75ex, parsep=0.75ex, wide, labelindent=0pt]
    \item We propose a distillation-free framework to generate high-quality training data, thereby enhancing end-to-end document conversion models.
    \item We propose a self-improvement method that effectively adapts document conversion models trained on synthetic data to real-world data distributions, without relying on external supervision.
    \item We develop a compact yet powerful document conversion model based on a public vision-language backbone, achieving state-of-the-art performance across various benchmarks and surpassing even some larger models.
\end{enumerate}

%% file: figures/figure_1.tex
\begin{figure*}[!htbp]
\centering
\includegraphics[width=\linewidth,scale=1.00]{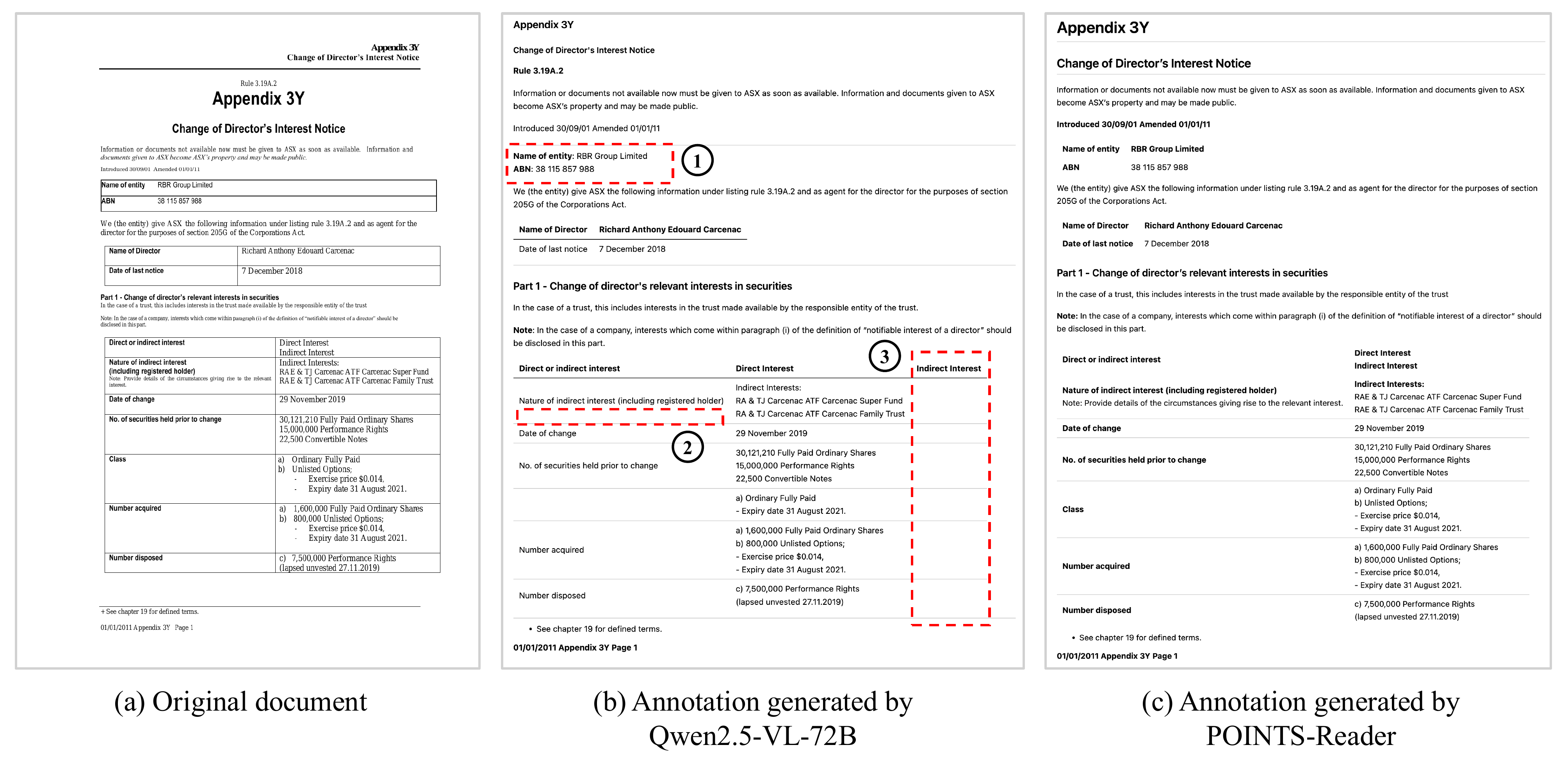}
\caption{\textbf{Example annotations generated by Qwen2.5-VL-72B and POINTS-Reader.} Distillation may not reach the performance of the teacher model and can inherit its biases, such as (1) failure to recognize tables, (2) missing text, and (3) incorrect table structures.}
\label{fig:fig1}
\end{figure*}

%% file: text/3-methods.tex
\section{Methods}
\label{sec:method}
\input{figures/figure_2}
This section presents our two-stage pipeline for constructing large quantities of high-quality data for document conversion tasks (\autoref{fig:fig2}). In Section~\ref{sec:warmup-stage}, we describe the creation of a large-scale, auto-rendered dataset with unified output formats for plain text, tables, and mathematical formulas, which is used to warm-up the model. In Section~\ref{sec:iterative-stage}, we detail our iterative self-improvement process, including data filtering strategies and their underlying motivations.

\subsection{Uniform Format Warm-up Stage}
\label{sec:warmup-stage}
\paragraph{Unified Output Format} Documents typically consist of four key elements: plain text, tables, mathematical formulas, and images. In this work, we focus on the first three—plain text (including body text, headers, footnotes, captions, etc.), tables, and mathematical formulas—each of which can be represented in various ways in Markdown (for example, tables may be written using native Markdown, HTML, or LaTeX syntax). Outputting these elements in heterogeneous formats complicates the learning process and may introduce confusion for the model. To facilitate effective model learning, we unify the output format for each of these key elements according to the following rules.

\begin{enumerate}[label={\bf \arabic*.}, leftmargin=*, topsep=0.5ex, itemsep=-0.5ex, partopsep=0.75ex, parsep=0.75ex, wide, labelindent=0pt]
\item \textbf{Plain Text:} Represented using Markdown syntax, following previous work\citep{lv2023kosmos}.
\item \textbf{Tables:} We adopt HTML syntax for table representation, as Markdown tables cannot handle complex structures (e.g., merged cells), and LaTeX tables lack standardization (their diverse syntax allows the same table to be represented in multiple ways). To further simplify, we remove all CSS except for merged cell attributes, and omit indentation and line breaks to reduce the token count.
\item \textbf{Mathematical Formulas:} Expressed in LaTeX\ syntax, following KaTeX\citep{KaTex} conventions: inline formulas are enclosed in single dollar signs (\$), and display formulas in double dollar signs (\$\$).
\end{enumerate}

Examples of these unified formats are provided in the appendix of ~\ref{sec:exam_unified}.

\paragraph{Large-scale Synthetic Data Generation}
To ensure that our data distribution closely resembles real-world scenarios, we generate data with a high degree of diversity. However, arranging and combining the aforementioned key document elements in various layouts results in an enormous number of possible configurations, making the data construction process highly complex. To address this, we aim to maximize data diversity while simultaneously simplifying the construction process. Consequently, we have created four categories of data: (1) plain text only, (2) text with mathematical formulas, (3) text with tables, and (4) multi-column layouts containing tables. The data generation process consists of two main steps. First, we design category-specific prompts and employ a large language model to generate the corresponding text (see~\ref{sec:prompt} for prompt details). Second, for tables and formulas, we apply rule-based filtering (the same as those described in the next section). The filtered data is then converted into HTML using templates for single-, two-, and three-column layouts, and rendered as images via Chrome’s headless mode. The resulting image-text pairs are used to fine-tune a general vision-language model, thereby enhancing its ability to extract and output document elements in a unified format.

\subsection{Iterative Self-improvement Stage}
\label{sec:iterative-stage}
While synthetic data enables large-scale training, its distribution, such as layout, often differs from that of real-world documents. To bridge this gap, we focus on acquiring high-quality real-world data. However, manual annotation is both costly and inefficient. Therefore, we leverage the model trained in the previous stage to generate annotations for real documents, making the quality of these generated annotations crucial to overall performance. To address this, we design a method that iteratively improves data quality through self-improvement. This approach, widely adopted in large language model development~\citep{grattafiori2024llama, yang2024qwen2, liu2024deepseekv3}, relies on effective data filtering strategies: only high-quality samples are retained for subsequent training. Our filtering strategies for plain text, tables, and formulas are rule-based, inspired by DeepSeek-R1~\citep{guo2025deepseekr1}, and are described below.

\paragraph{Filtering Plain Text}
The primary challenges in visual text extraction are hallucinations, repetition, and omissions, particularly when dealing with complex layouts~\citep{nassar2025smoldocling}. Following the approach of CCOCR~\citep{yang2024cc}, we employ the F1-score to filter plain text. Specifically, we extract reference text using a traditional OCR model (e.g., PaddleOCR~\citep{du2020pp}), and normalize both the model predictions and references by: (1) removing all non-alphanumeric characters and splitting the text by spaces, and (2) counting the occurrences of each unit. We denote the statistics for the model prediction and reference as $P = \{(u_p^0, c_p^0), (u_p^1, c_p^1), \ldots, (u_p^{N-1}, c_p^{N-1})\}$ and $T = \{(u_t^0, c_t^0), (u_t^1, c_t^1), \ldots, (u_t^{N-1}, c_t^{N-1})\}$, respectively, where $u_x^i$ represents a basic unit (e.g., a single word) and $c_x^i$ its occurrence count. Precision, recall, and F1-score are then computed as follows:

\begin{equation}
    \text{Precision} = \frac{\sum_{i=0}^{N-1}\min(c_p^i, c_t^i)}{\sum_{i=0}^{N-1}c_p^i}
\end{equation}

\begin{equation}
    \text{Recall} = \frac{\sum_{i=0}^{N-1}\min(c_p^i, c_t^i)}{\sum_{i=0}^{N-1}c_t^i}
\end{equation}

\begin{equation}
    \text{F1} = \frac{2 \times \text{Precision} \times \text{Recall}}{\text{Precision} + \text{Recall}}
\end{equation}

Samples with F1-scores below a threshold (e.g., 0.9) are discarded. As traditional OCR models are robust to the aforementioned issues, such as missing main parts of main text and hallucinations, outliers in filtered data are significantly reduced.
\paragraph{Filtering Tables}
Existing table structure recognition models (e.g., SLANet~\citep{li2022slanet}, StructEqTable~\citep{xia2024docgenome}) tend to be less robust and are often restricted to images containing only tables. Consequently, we do not use their predictions as reference answers. Instead, we focus on ensuring table structural validity: for each table in the model output, we verify the consistency of the number of cells in each row and column. Samples with invalid table structures are subsequently removed.

\paragraph{Filtering Mathematical Formulas}
For formulas, we verify only syntactic correctness, not semantic validity. All formulas are extracted from the model output and checked for syntax errors. Samples containing invalid formulas are discarded.

All samples that pass these filters are used to retrain the model. This process is repeated for several rounds, resulting in significant improvements in both model performance and data quality. Although we do not verify the content of tables and mathematical formulas, the recognition accuracies for these two elements also steadily improve during this stage (see \autoref{fig:fig8}).

%% file: figures/figure_2.tex
\begin{figure*}[!htbp]
\centering
\includegraphics[width=\linewidth,scale=1.00]{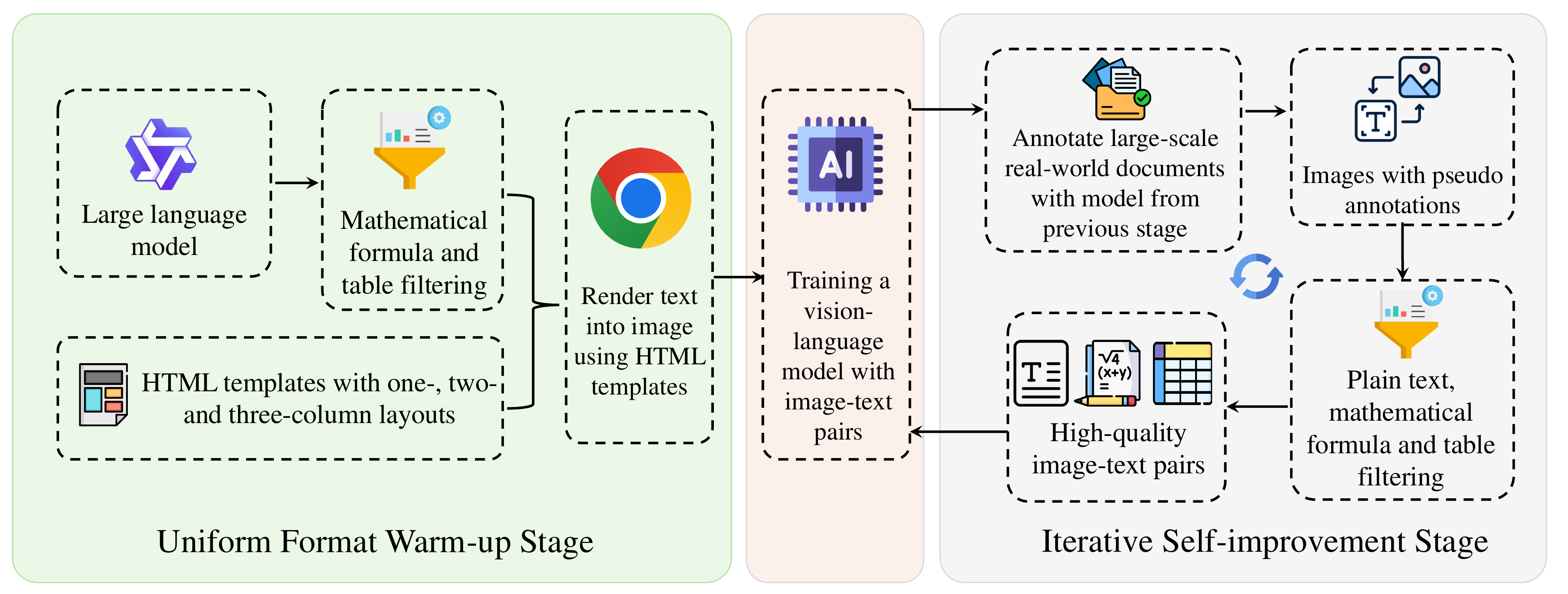}
\caption{\textbf{Demonstration of the two-stage pipeline to generate large-scale high quality dataset.}}
\label{fig:fig2}
\end{figure*}

%% file: text/4-exps.tex
\section{Experiments}
This section is organized into three main parts. In Section~\ref{sec:exp-settings}, we detail the experimental settings, including the datasets, model architecture, and other relevant implementation details. Next, in Section~\ref{sec:abla}, we present comprehensive ablation studies to evaluate the contribution of each component in our design and highlight several noteworthy findings. Finally, in Section~\ref{sec:comparison}, we compare our model against state-of-the-art methods across a range of benchmark datasets.

\subsection{Experiment Settings}
\label{sec:exp-settings}
\paragraph{Datasets}
During the uniform format warm-up stage, we observed that prompting the LLM to generate text containing tables typically resulted in simple table structures, rarely featuring merged cells or complex layouts. To introduce greater structural diversity, we selected a subset of tables from the PubTabNet~\citep{zhong2020image} training set and prompted the LLM to generate descriptive paragraphs based on their content. The corresponding tables were then randomly inserted into the generated text to create more realistic and challenging samples. For the iterative self-improvement stage, we adopted DocMatix~\citep{laurençon2024building} as our primary dataset. DocMatix, curated from PDFA, contains over two million document images spanning a wide range of scenarios, including academic papers and various document types. In each iteration, our model was used to perform inference on DocMatix, after which the results were filtered and the high-quality outputs were used to further train the model of the next version.

\paragraph{Model Training}
We used POINTS-1.5~\citep{liu2024points1} as our base model and Qwen2.5-3B-Instuct~\citep{yang2024qwen2} as the large language model (LLM) to balance efficiency and effectiveness. Following the POINTS-1.5 training paradigm, our approach consists of two stages: pre-training and visual instruction tuning. The pre-training stage uses the same data as POINTS-1.5, while the visual instruction tuning stage incorporates all newly generated data from both the warm-up and self-improvement stages. Additionally, we included the general datasets used in POINTS-1.5 to further enhance model performance. All training hyperparameters and other settings, except for the data used in visual instruction tuning and maximum context length (8192 in this work) , are kept identical to those in POINTS-1.5.

\paragraph{Evaluation}
To comprehensively assess our model’s extraction capabilities across plain text, tables, and mathematical formulas, we employ two benchmark datasets: English split of Fox~\citep{liu2024focus} and English split of  OmniDocBench~\citep{ouyang2024omnidocbenchbenchmarkingdiversepdf}. More details about the two benchmarks can be found in the appendix.

\subsection{Ablation Studies and Observations}
\label{sec:abla}
In this section, we present comprehensive experiments to evaluate the effectiveness of each component in our two-stage data construction pipeline. Unless otherwise noted, all results are reported using the overall score from OmniDocBench~\citep{ouyang2024omnidocbenchbenchmarkingdiversepdf}, which assesses performance across plain text, tables, and formulas. 

\input{tables/table_1}

\subsubsection{Uniform Format Warm-up Stage}
\label{sec:abla_uws}
\input{figures/figure_3}
\paragraph{Increasing data diversity leads to improved experimental results.} 
As previously discussed, we constructed four types of data for this stage: (1) plain text, (2) text containing mathematical formulas, (3) text with tables, and (4) multi-column layouts with tables. For each type, we generated 200{,}000 samples (after LaTeX formula and table filtering). Using the SFT data from POINTS-1.5 as a baseline, we incrementally incorporated these new data types. The detailed results are presented in \autoref{tab:tab1}. The findings indicate that adding data from a specific category not only enhances the model’s performance on that category, but also leads to progressive improvements in overall metrics. Notably, the inclusion of multi-column layout data (multi-column layouts with tables) resulted in a substantial boost in reading order performance.

\paragraph{When the dataset reaches a certain scale, the model’s performance plateaus.} To investigate this, we linearly increased the number of samples per category, expanding the total dataset size from 100,000 to 1.2 million (\autoref{fig:fig3}(a)). As shown, model performance ceases to improve—and even declines—once the data size exceeds 800,000. This decline may be attributed to the fact that the distribution of these synthetic data, such as layout, differs substantially from real-world data. As a result, further increasing the volume of such synthetic data can lead to overfitting to these artificial patterns, ultimately degrading the model’s generalization ability. These findings underscore the necessity of the \textbf{iterative self-improvement stage}.

\input{tables/table_2}

\paragraph{Excluding data with abnormal aspect ratios can enhance model performance.}
After obtaining these synthetic data, we manually inspected a subset of images and found that some exhibited abnormal shapes, such as being excessively long. To further improve model performance, we investigated the impact of filtering out samples with abnormal aspect ratios generated during the Uniform Format Warm-up Stage. Since A4 paper—the most prevalent document layout—has an aspect ratio of $\sqrt{2}$ (ISO 216), we first analyzed the distribution of aspect ratios in our rendered dataset. We then conducted ablation studies to determine the optimal range for filtering. As shown in \autoref{tab:tab2}, removing images with aspect ratios outside the range $(\frac{2}{5}, \frac{5}{2})$ leads to improvements in model accuracy. \autoref{fig:fig3}(b) further visualizes the aspect ratio distribution before and after filtering (i.e., samples between the red dotted lines). 

\subsubsection{The Iterative Self-improvement Stage}
\label{sec:abla_iss}
\paragraph{These filtering strategies effectively increase the quality of the data.} 

We first perform full inference on DocMatrix~\citep{laurençon2024building} using the model obtained from the Uniform Format Warm-up Stage. Subsequently, we sequentially apply our proposed filtering methods for plain text, tables, and mathematical formulas to the dataset. The filtered data is then used for further model training. Specifically, for plain text, we discard all samples with an F1-score below 0.9. The performance of the resulting models is summarized in~\autoref{tab:tab3}. As shown, the model’s performance consistently improves with the application of additional filtering strategies. Moreover, each filtering method plays a crucial role in enhancing the model’s effectiveness on the corresponding tasks. For example, applying the plain text filter reduces the edit distance on text-related metrics from 0.470 to 0.380. Additionally, we observe that even though the baseline (without any filtering strategies) utilizes 2 million real-world samples from DocMatix, it yields only marginal improvement compared to the model trained solely on synthetic data. This phenomenon underscores the importance of data quality.

\input{tables/table_3}
\paragraph{The F1-score threshold plays a crucial role in ensuring the quality of training data.} Setting the threshold too low introduces excessive low-quality data, which can hinder model performance. Conversely, setting the threshold too high risks discarding a substantial amount of otherwise useful data, thereby reducing the diversity of the training set and negatively impacting model training. This is because traditional OCR models are not flawless (unable to recognize formulas) and some predictions from our model may only omit minor elements, such as page numbers or headers, while still containing all of the main content. To investigate this, we conducted a comprehensive ablation study on different F1-score thresholds, as shown in \autoref{tab:tab4}. The results indicate that both overly low and overly high thresholds are detrimental to model performance.
\input{tables/table_4}
\input{figures/figure_4-1}

\paragraph{We observe improved quality of the real-world data to train our model.} As illustrated in \autoref{fig:fig4-1}, increasing the number of iterations consistently enhances model performance. This improvement is also evident in the quality of the generated data, which benefits from additional iterations. To quantitatively assess this trend, we calculate the F1-score between the model’s predictions on DocMatrix images and the results produced by PaddleOCR, averaging these scores across all images (\autoref{fig:fig4-2}). The results demonstrate that, as the number of iterations increases, the model’s predictions become increasingly aligned with the target outputs. Prior to each training cycle, we apply the three previously described data filtering methods. The amount of data retained after filtering serves as an additional indicator of data quality. As shown in \autoref{fig:fig4}, the number of samples containing all three key elements rises with more iterations. However, although the model continues to improve, the rate of progress has begun to slow, suggesting that performance may soon plateau. To achieve further improvements, it will be necessary to explore additional strategies, such as increasing data diversity, which we leave for future work.

\input{tables/table_5}
\input{figures/figure_4-2}

\input{figures/figure_4}

\subsection{Comparison with Other Models} 
We compare our model against three categories of baselines: pipeline methods, general vision-language models, and expert OCR models. For each category, we select the most representative approaches in the industry, with detailed results presented in \autoref{tab:tab5}. To comprehensively assess performance on plain text, mathematical formulas, and tables, we evaluate all models on four benchmarks. As shown in the table, both general vision-language models and specialized OCR models still lag behind pipeline methods, indicating that end-to-end approaches have considerable room for improvement. Compared to general vision-language models such as Qwen2.5-VL-72B, POINTS-Reader matches or even surpasses larger models on several benchmarks; for instance, it outperforms Qwen2.5-VL-72B on the table metric of OmniDocBench and PubTabNet. Against expert OCR models, POINTS-Reader surpasses the proprietary Mistral OCR by a noticeable margin. Notably, POINTS-Reader excels in table recognition, outperforming GOT-OCR by 0.197 on the Table metric of OmniDocBench.
\label{sec:comparison}

%% file: tables/table_1.tex
\begin{table}[ht]
\centering
\tabcolsep 1.5pt
\scalebox{0.9}{
\begin{tabular}{lccccc}
\rowcolor{gray!30}
Method&Text$\downarrow$&Table$\downarrow$&Formula$\downarrow$& Order$\downarrow$&Overall$\downarrow$\\
\hline
Baseline & 0.551 & 0.652 & 0.730 & 0.570 & 0.626 \\
+Text & 0.522 & 0.641 & 0.721 & 0.553 & 0.609 \\
+Formula & 0.513 & 0.640 & 0.600 & 0.530 & 0.571 \\
+Table & 0.495 & 0.590 & 0.595 & 0.523 & 0.551 \\
\rowcolor{green!10}
+Multi-Column & 0.485 & 0.572 & 0.511 & 0.471 & 0.510 \\
\end{tabular}
}
\vspace{-0.5em}
\caption{\textbf{Increasing the diversity of data leads to improved performance (lower is better).} baseline: model fine-tuned with visual instruction tuning data from POINTS-1.5\citep{liu2024points1}. Order: reading order performance. Data is incrementally added from each category.}
\label{tab:tab1}
\end{table}

%% file: figures/figure_3.tex
\begin{figure*}[!htbp]
\centering
\includegraphics[width=\linewidth,scale=1.00]{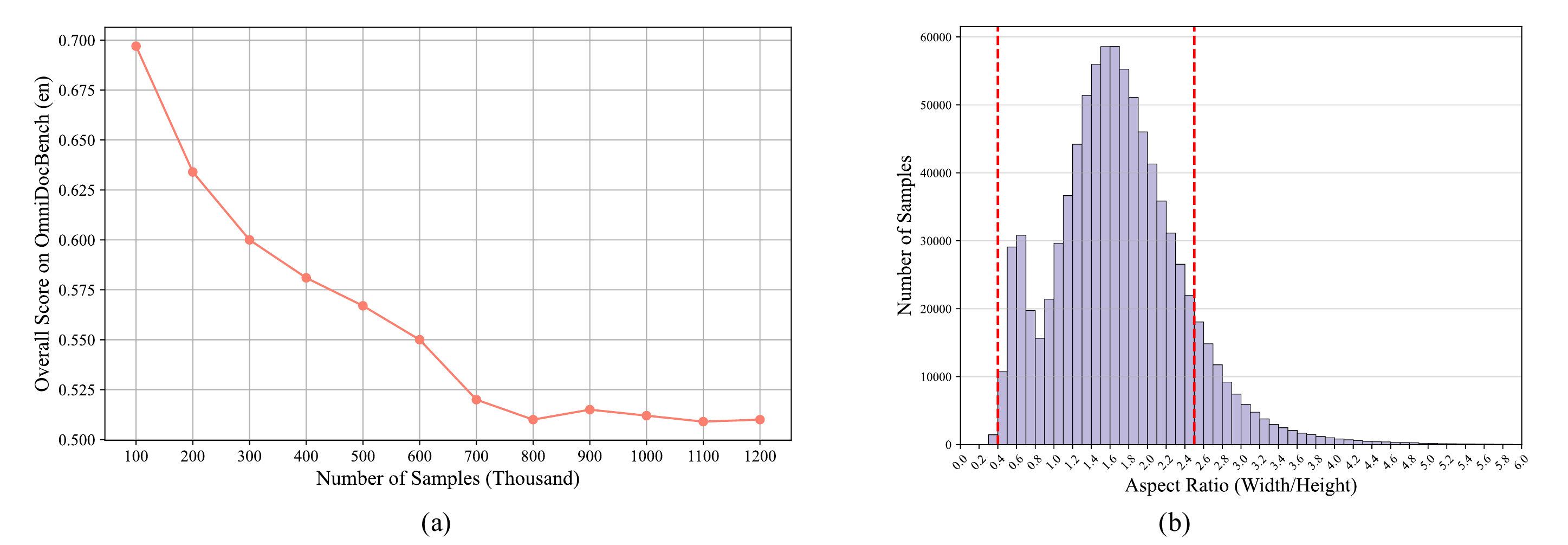}
\vspace{-0.8cm}
\caption{(a) Scaling curve of data generated during the uniform format warm-up stage (lower is better). (b) Distribution of aspect ratios (width/height) in the original dataset. Samples with aspect ratios beyond the red dotted line are filtered out.}
\label{fig:fig3}
\end{figure*}

%% file: tables/table_2.tex
\begin{table}[ht]
\centering
\tabcolsep 1.5pt
\scalebox{0.95}{
\begin{tabular}{lccccc}
\rowcolor{gray!30}
Range&Text$\downarrow$&Table$\downarrow$&Formula$\downarrow$& Order$\downarrow$&Overall$\downarrow$\\
\hline
$(0, +\infty)$ & 0.485 & 0.572 & 0.511 & 0.471 & 0.510 \\
$(\frac{1}{5}, 5)$ & 0.483 & 0.570 & 0.520 & 0.455 & 0.507 \\
$(\frac{1}{4}, 4)$ & 0.480 & 0.571 & 0.519 & 0.453 & 0.506 \\
$(\frac{1}{3}, 3)$ & 0.480 & 0.585 & 0.513 & 0.434 & 0.503 \\
\rowcolor{green!10}

$(\frac{2}{5}, \frac{5}{2})$ & 0.475 & 0.572 & 0.522 & 0.421 & 0.498 \\
$(\frac{2}{3}, \frac{3}{2})$ & 0.479 & 0.574 & 0.524 & 0.430 & 0.502 \\
\end{tabular}
}
\vspace{-0.5em}
\caption{\textbf{Restricting the aspect ratio of image (automatically generated in the Uniform Format Warm-up Stage) within a range can boost the performance (lower is better).} Range: we only keep images whose aspect ratio (heigh \vs width) are in within this range.}
\label{tab:tab2}
\end{table}

%% file: tables/table_3.tex
\begin{table}[ht]
\centering
\tabcolsep 1.5pt
\scalebox{0.85}{
\begin{tabular}{lccccc}
\rowcolor{gray!30}
Filtering Strategy&Text$\downarrow$&Table$\downarrow$&Formula$\downarrow$& Order$\downarrow$&Overall$\downarrow$\\
\hline
N/A & 0.470 & 0.561 & 0.514 & 0.430 & 0.493 \\
+Text & 0.380 & 0.551 & 0.501 & 0.418 & 0.463 \\
+Table & 0.378 & 0.494 & 0.500 & 0.414 & 0.447 \\
\rowcolor{green!10}
+Formula & 0.374 & 0.492 & 0.457 & 0.434 & 0.439 \\
\end{tabular}
}
\vspace{-0.5em}
\caption{\textbf{Rule-based data filtering strategies significantly enhance model performance}. ``N/A'' indicates that no filtering strategies are applied; all data generated by the model during the uniform format warm-up stage are used for training in this stage.} 
\label{tab:tab3}
\vspace{-0.3cm}
\end{table}

%% file: tables/table_4.tex
\begin{table}[ht]
\centering
\tabcolsep 1.5pt
\scalebox{0.90}{
\begin{tabular}{lccccc}
\rowcolor{gray!30}
F1-threshold&Text$\downarrow$&Table$\downarrow$&Formula$\downarrow$& Order$\downarrow$&Overall$\downarrow$\\
\hline
0.70 & 0.399 & 0.505 & 0.469 & 0.455 & 0.457 \\
0.80 & 0.387 & 0.504 & 0.466 & 0.451 & 0.452 \\
\rowcolor{green!10}
0.90 & 0.374 & 0.492 & 0.457 & 0.434 & 0.439 \\
0.95 & 0.381 & 0.496 & 0.460 & 0.438 & 0.444 \\
\end{tabular}
}
\vspace{-0.5em}
\caption{\textbf{The threshold of F1-score for plain-text filtering is important.}} 
\label{tab:tab4}
\vspace{-0.3cm}
\end{table}

%% file: figures/figure_4-1.tex
\begin{figure}[!htbp]
\centering
\includegraphics[width=\linewidth,scale=1.00]{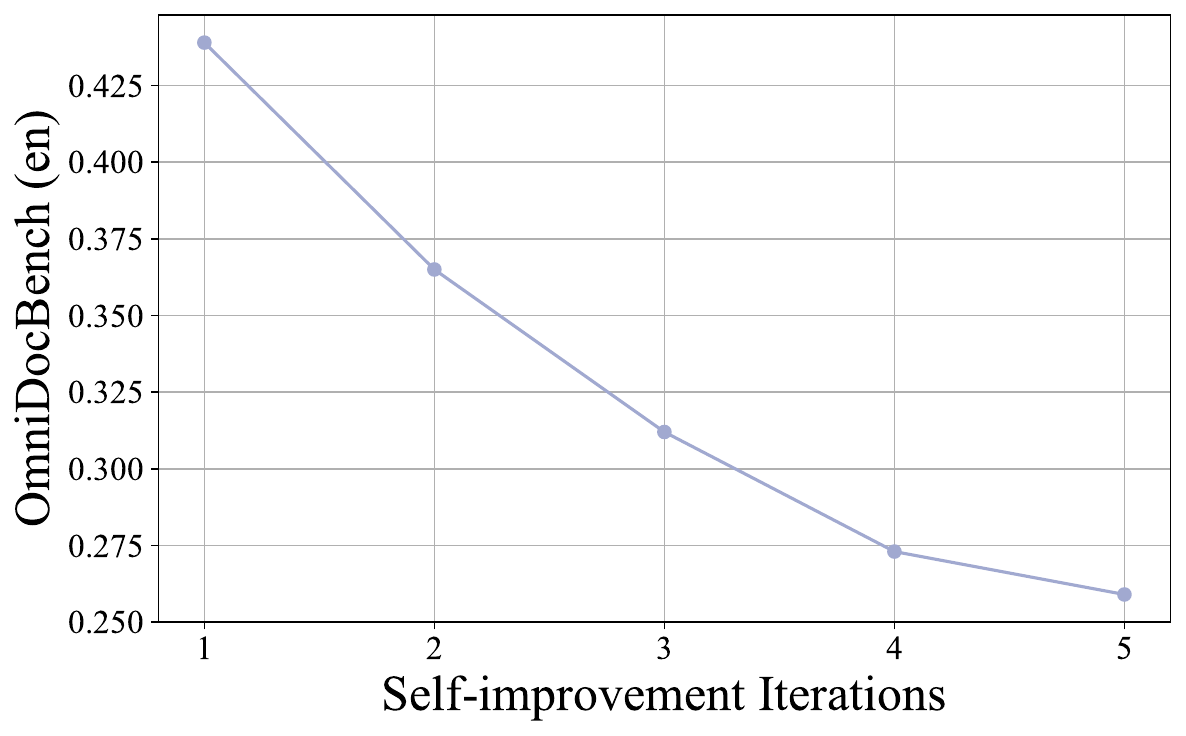}
\caption{\textbf{Model performance steady improves during the self-improvement stage.}}
\label{fig:fig4-1}
\vspace{-0.5cm}
\end{figure}

%% file: tables/table_5.tex
\begin{table*}[!ht]
\centering
\tabcolsep 7pt
\scalebox{0.7}{
\begin{tabular}{llcccccc}
 \multicolumn{2}{c}{Evaluation Protocol$\rightarrow$} & \multicolumn{5}{c}{OmniDocBench} & Fox \\
\cmidrule(lr){3-7} \cmidrule(lr){8-8}
Method&Size& Text$\downarrow$ & Formula$\downarrow$ & Tabl$\downarrow$ & Order$\downarrow$& Overall$\downarrow$& Edit Dist$\downarrow$\\
\midrule
\multicolumn{7}{@{\;}l}{\bf Pipeline Methods} \\
MinerU\cite{wang2024mineru} & - & 0.061 & 0.278 & 0.18 & 0.079 & 0.150 & - \\
Marker\cite{marker} & - & 0.080 & 0.530 & 0.619 & 0.114 & 0.336 & - \\
Mathpix\citep{mathpix} & - & 0.105 & 0.306 & 0.243 & 0.108 & 0.191 & - \\
\midrule
\multicolumn{7}{@{\;}l}{\bf General Vision-language Model} \\
Qwen2.5-VL-3B\citep{bai2025qwen2} & 3B & 0.252 & 0.429 & 0.612 & 0.268 & 0.390 & 0.063 \\
Qwen2.5-VL-7B\citep{bai2025qwen2} & 7B & 0.144 & 0.436 & 0.590 & 0.154 & 0.331 & 0.032 \\
Qwen2.5-VL-72B\citep{bai2025qwen2} & 72B & 0.092 & 0.315 & 0.341 & 0.106 & 0.214 & 0.027 \\
\midrule
\multicolumn{7}{@{\;}l}{\bf Expert Vision-language Model} \\
GOT-OCR\citep{wei2024general} & 716M & 0.189 & 0.360 & 0.532 & 0.141 & 0.287 & 0.035 \\
Nougat\citep{blecher2023nougat} & 350M & 0.365 & 0.488 & 0.572 & 0.382 & 0.452 & - \\
Mistral OCR	& - & 0.072 & 0.318 & 0.600 & 0.083 & 0.268 & - \\
OLMOCR\citep{poznanski2025olmocr} & 7B & 0.097 & 0.455 & 0.608 & 0.145 & 0.326 & - \\
POINTS-Reader & 3B & 0.176 & 0.383 & 0.335 & 0.144 & 0.259 & 0.023 \\

\end{tabular}
}
\caption{\textbf{Comparison with other methods (pipeline and end-to-end) across four benchmarks.} The performance of the Qwen2.5-VL series is reported using the same evaluation settings as POINTS-Reader. For other methods, we use the metrics reported in their original papers, or, when unavailable, from subsequent works.}
\label{tab:tab5}
\end{table*}

%% file: figures/figure_4-2.tex
\begin{figure}[!htbp]
\centering
\includegraphics[width=\linewidth,scale=1.00]{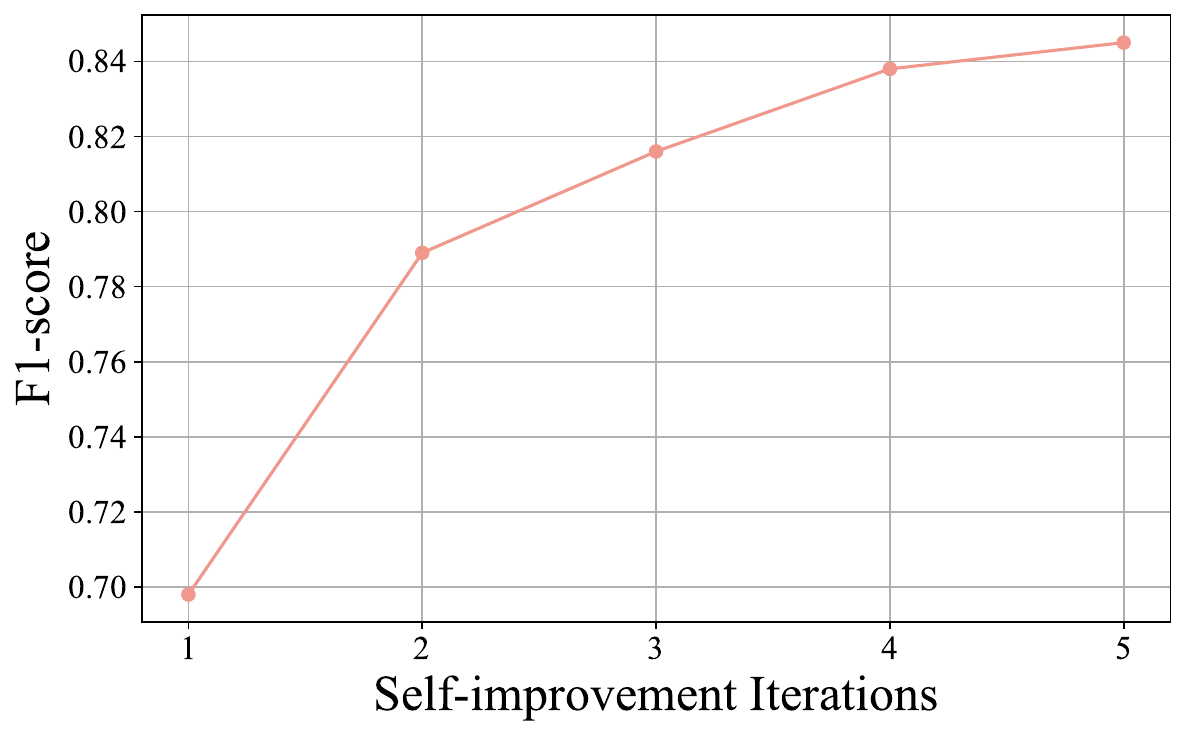}
\caption{\textbf{The F1-score steadily improves during the self-improvement stage.} The score is computed prior to data filtering.}
\label{fig:fig4-2}
\vspace{-0.4cm}
\end{figure}

%% file: figures/figure_4.tex
\begin{figure*}[!htbp]
\centering
\includegraphics[width=\linewidth,scale=1.00]{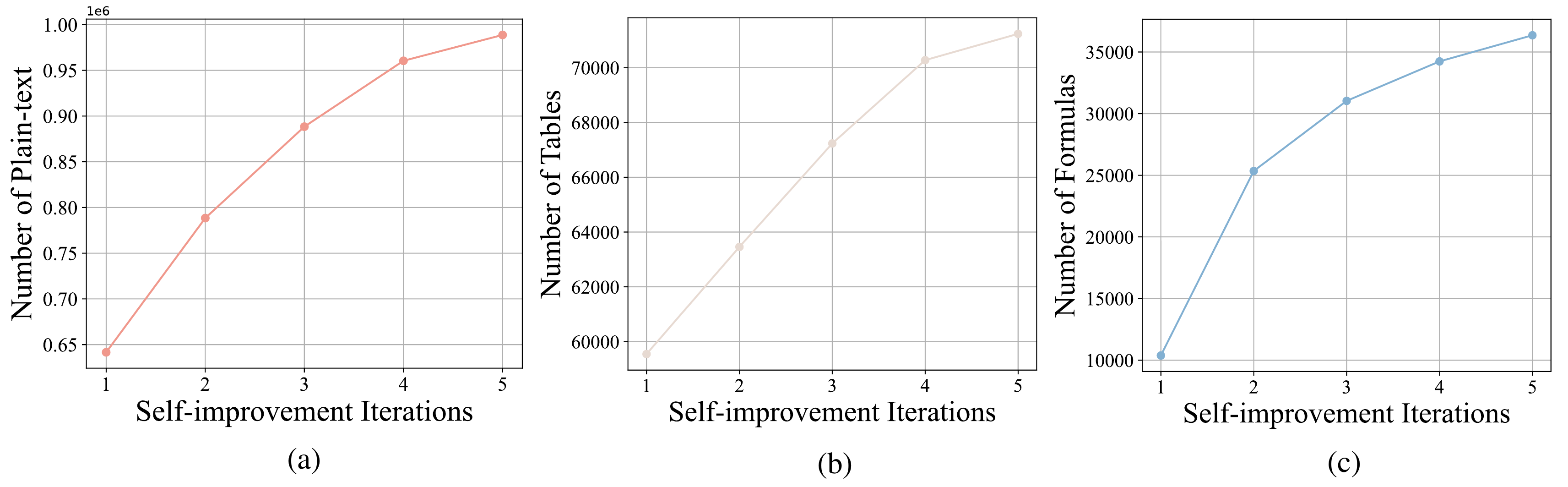}
\vspace{-0.8cm}
\caption{\textbf{The number of samples after filtering consistently increases.}(a) The number of retained samples containing only plain text increases after filtering. (b) The number of retained samples containing tables increases after filtering. (c) The number of retained samples containing tables increases after filtering.}
\label{fig:fig4}
\end{figure*}

%% file: text/2-related.tex
\section{Related Works}
\paragraph{Vision-Language Models}
Recent vision-language models~\citep{liu2024improved, liu2024points, liu2024points1, liu2023visual, liu2024rethinking, zhang2023internlm, li2024llava, bai2025qwen2, chen2024expanding} have made significant advances. BLIP2~\citep{li2023blip} reduced training cost by only updating the Q-Former. LLaVA~\citep{liu2023visual} further simplified modality alignment and introduced large-scale instruction tuning. Later works~\citep{chen2024expanding, zhang2024internlm} enabled flexible image resolutions via tiling, while Qwen2-VL~\citep{wang2024qwen2} and Qwen2.5-VL~\citep{bai2025qwen2} adopted NaViT-style~\citep{dehghani2023patch} encoders to natively support arbitrary resolutions. Improved evaluation benchmarks~\citep{liu2024mmbench} have also accelerated progress.

\paragraph{Document Reading}
Previously, a series of excellent works \citep{lee2023pix2struct, hu-etal-2024-mplug, hu2024mplug2.0} have aimed to optimize the performance of document conversion and comprehension. Recently, document conversion methods have primarily been categorized into \textbf{pipeline approaches}, which utilize specialized models and extensive manual processing \citep{wang2024mineru, marker, Docling}, and \textbf{end-to-end approaches}, which involve training vision-language models directly on document datasets \citep{wei2024general, blecher2023nougat, poznanski2025olmocr}. However, existing datasets often suffer from noise or rely on distillation from large models. To address these issues, we propose an automatic pipeline to construct large, high-quality datasets for end-to-end training without the need for distillation.

%% file: text/5-con.tex
\section{Conclusion}
Traditionally, end-to-end document conversion models are developed by distilling knowledge from proprietary models such as GPT-4o or large-scale open-source models like Qwen2-VL-72B. However, distillation-based approaches face several challenges, including limited scientific innovation and the risk of inheriting the shortcomings of the source models. In this paper, we propose a novel two-stage framework for generating large-scale, high-quality training data for end-to-end document conversion without relying on model distillation. In the first stage, we automatically construct a large dataset and train a model to produce unified outputs for diverse document elements. This model is then employed to generate image-text pairs from real documents, which are rigorously filtered using carefully designed strategies. The resulting high-quality data is subsequently used to further train the model. By iteratively repeating this process, we achieve substantial improvements in both data quality and model performance. Our approach ultimately yields a model, achieving state-of-the-art performance across various benchmarks and surpassing even some larger models.

\section{Limitations}
Currently, our model supports only English, which limits its applicability in other widely used languages, such as Chinese and Japanese. Additionally, all datasets used in our current experiments consist of printed fonts, resulting in suboptimal performance when processing handwritten text, such as notes. In the future, we plan to continuously enhance the multilingual capabilities of \textbf{POINTS-Reader} and improve its support for handwritten fonts. Moreover, at present, our model only supports the extraction of plain text, formulas, and tables. We aim to extend this functionality to include the extraction of images, such as identifying and outputting their locations within documents.

%% file: text/6-appendix.tex
\section{Appendix}
\label{sec:appendix}
\subsection{More Experiment Results}
\paragraph{Loading weights from a previous model can degrade performance.} We compare the performance of models initialized with weights from a pre-trained model and those initialized with weights from a previous version of the model. As shown in \autoref{fig:fig5}, models initialized from the pre-trained model consistently outperform those initialized from the previous version. The data used in the previous version inherently contains some noise, which can negatively affect the model trained on it. In contrast, models initialized from the pre-trained model are not subject to this issue.
\input{figures/figure_5}

\paragraph{Including data generated in the UWS into the ISS will benefits the performance of the model.} We examine the effect of integrating data generated in UWS into ISS on model performance. The figure below compares model performance before and after the inclusion of UWS-generated data. The results indicate that incorporating data from UWS positively influences model performance. While UWS-generated data is somewhat limited in diversity, its annotations are highly accurate. Consequently, introducing this data into ISS complements the data produced during the ISS stage and further improves the model’s performance.
\input{figures/figure_6}

\input{figures/figure_7}
\paragraph{Comparison with direct distillation from Qwen2.5-VL-72B.} In this section, we compare our model with a counterpart trained on data distilled from Qwen2.5-VL-72B~\cite{bai2025qwen2}. As shown in \autoref{tab:tab6}, our model—without any distillation—outperforms the model distilled from Qwen2.5-VL-72B by a significant margin. Although Qwen2.5-VL-72B achieves higher overall performance on OmniDocBench compared to PONTS-Reader, the distillation strategy introduces several challenges: (1) it introduces uncertainty in the performance of the student model, which often struggles to fully match the capabilities of the teacher model; and (2) distilling from large-scale flagship models, particularly on sizable datasets, imposes a substantial computational burden.

\input{tables/table_6}

\paragraph{Distribution of the data in the final iteration of the Self-improvement Stage.} The figure above illustrates the data distribution in the final iteration of the self-improvement stage. In total, we utilized 2,234,134 DocMatrix images, of which 1,096,325 were used for training in the last stage. \autoref{fig:fig7}~(a) presents the distribution of samples that (1) contain only plain text, (2) contain tables, and (3) contain formulas, with 0.1\% of the samples containing both tables and formulas. It is evident that samples containing only plain text constitute the vast majority, while those containing tables or formulas represent a much smaller proportion. \autoref{fig:fig7}~(b) displays the distribution of sample counts with respect to different token lengths. Most samples have a length of fewer than 1,000 tokens.

\paragraph{Data balance.} As illustrated in \autoref{fig:fig7}~(a), there is a significant imbalance among samples containing plain text, tables, and formulas. To address this issue, we balance these sample types during training and conduct extensive experiments to determine the optimal sampling ratios for these three key elements. The detailed results are presented in \autoref{tab:tab7}. As shown, down-sampling plain text samples and up-sampling table and formula samples do not lead to improvements; in fact, they often result in inferior performance. We hypothesize that this is primarily because down-sampling plain text reduces the diversity of training samples, while up-sampling tables and formulas does not increase diversity. Consequently, these approaches impair the generalization ability of the trained model.
\input{tables/table_7}

\paragraph{Steady improvement}
\input{figures/figure_8}
As shown in \autoref{fig:fig8}, the model's performance on plain text, tables, and formulas exhibits steady improvement during the self-improvement stage. Notably, although we employ only rule-based filtering strategies—verifying solely the structural integrity of tables and the syntactic correctness of mathematical formulas, without assessing the actual content correctness—the model still demonstrates consistent improvement in recognizing these two elements. These results further demonstrate the effectiveness of such filtering strategies in selecting high-quality data, and highlight a promising direction for future research.

\paragraph{Further analysis of improvements during the self-improvement stage.}
We first examine how many samples from the previous iteration's training set are retained, and plot the corresponding ratios on the left side of the following figure. Next, we measure the F1-score of these retained samples and plot the average F1-score statistics on the right side of the figure (for the first iteration, the F1-score is computed across all samples). Combined with the results from \autoref{fig:fig4}, we hypothesize that the observed improvements arise from two factors: (1) enhanced quality of the original training data, and (2) increased data diversity resulting from the inclusion of more filtered samples.
\input{figures/figure_12}

\subsection{Comparison with Other Datasets}
Although the primary objective of our work is to propose an automated approach for generating high-quality data to train end-to-end document conversion models, we also compare the dataset produced in the final iteration of our self-improvement process with those used in previous studies from multiple perspectives. Currently, open-source datasets that meet the following criteria are extremely scarce: (1) they contain plain text, tables, and mathematical formulas simultaneously; and (2) their table annotation format can fully represent the structure of all tables, such as through HTML. \autoref{tab:tab8} provides a comparison between our dataset and those utilized in related works. It can be observed that the two currently available datasets for end-to-end training both utilize Markdown as the table representation format. However, as previously discussed, Markdown offers limited expressiveness for representing complex tables. Furthermore, both datasets employ distillation methods during their data construction processes. For instance, KOSMOS leverages the Microsoft Read API to annotate scanned documents, while olmOCR relies entirely on data distilled from GPT-4o. Although KOSMOS-2.5 contains a substantial amount of data, its construction approach has certain limitations, such as representing tables in Markdown and directly parsing content from PDFs, which can introduce issues—particularly with mathematical formulas. Consequently, the annotation quality of these datasets is also limited.
\input{tables/table_8}

\subsection{Case Study of Samples Evolved During Iterations}
We randomly select three samples generated by the model in both the first and last iterations. The comparisons are presented in the following figures. As shown, the quality of the annotations improves significantly as the model's performance increases.

\subsection{More Details about Evaluation}
\paragraph{Fox}: We utilized the English evaluation split, Fox-Page-en, from Fox to assess end-to-end page conversion. Fox-Page-en comprises 112 English pages, featuring both single-column and double-column layouts. Each page contains over 1,000 words, making it a challenging testbed for document image parsing. The evaluation metrics primarily measure the Normalized Edit Distance between the model's output and the target.

\paragraph{OmniDocBench}: OmniDocBench is primarily used for evaluating end-to-end document conversion, encompassing 19 layout types. It includes assessments of text, formulas, tables, and output reading order. Our evaluation and annotation schemes are consistent, with metrics based on Normalized Edit Distance.

\onecolumn
\input{figures/figure_9}
\input{figures/figure_10}
\input{figures/figure_11}

\subsection{Computational Analysis} 

We conduct all experiments using 64 Nvidia H800 GPUs. Training on 1 million data samples takes approximately 7 hours. In addition we deploy the model with SGLang, and inference on 2 million DocMatrix samples requires about 10 hours.

\subsection{Prompts For Large-scale Synthetic Data Generation in the Uniform Format Warm-up Stage}
\label{sec:prompt}

\definecolor{darkorange}{RGB}{255, 140, 0}
\definecolor{darkblue}{RGB}{84, 112, 198}
\definecolor{lightgreen}{RGB}{145, 204, 117}
\definecolor{lightyellow}{RGB}{250, 200, 88}
\definecolor{lightred}{RGB}{238, 102, 102}
\definecolor{lightblue}{RGB}{115, 192, 222}

\newtcolorbox{promptbox}[2][Prompt]{
colback=black!5!white,
arc=5pt, 
boxrule=0.5pt,
fonttitle=\bfseries,
breakable,
title=#1, 
before upper={\small}, fontupper=\fontfamily{ptm}\selectfont,
colframe=#2, 
}

\begin{promptbox}[Prompt for Plain Text]{lightgreen}
Please select one of the following topics and write a MarkDown text with the following requirements with a random seed:\\
\\
1.\hspace{0.5em}Choose "TOPIC" as the topic with a word count of approximately 300-500 words.\\
2.\hspace{0.5em}The language and formatting style of the text should be chosen from the provided styles below.\\
3.\hspace{0.5em}Do not include any tables or mathematical formulas in the text.\\
4.\hspace{0.5em}You may choose to use some of the following MarkDown syntax elements in your writing:\\
   \hspace*{1em} - Different levels of headings\\
   \hspace*{1em} - Bold\\
   \hspace*{1em} - Italics\\
   \hspace*{1em} - Bold italics\\
   \hspace*{1em} - Underline\\
   \hspace*{1em} - Superscript\\
   \hspace*{1em} - Subscript\\
   \hspace*{1em} - Lists\\
     \hspace*{2em} - Unordered lists\\
     \hspace*{2em} - Ordered lists\\
5.\hspace{0.5em}The overall style and organization of the text should be more varied, **avoid always adding conclusions or summaries at the end**.\\
6.\hspace{0.5em}The content of the text does not have to be complete; it can be an excerpt.\\
7.\hspace{0.5em}Only return the generated text content without any additional explanations, such as descriptions before or after the text.\\
8.\hspace{0.5em}Do not mention the selected language, or formatting style in the text.\\
9.\hspace{0.5em}Please provide the response in English.\\
10.\hspace{0.5em}Random seed is SEED\\

Available text languages and formatting styles:\\
Exam paper, slides, academic paper, book, textbook, magazine, notes, newspaper, financial report
\end{promptbox}

\begin{promptbox}[Prompt for Formulas]{lightgreen}
Please choose one of the following topics and write a MarkDown text with the following requirements with a random seed:\\

1.\hspace{0.5em}Choose "TOPIC" as the topic and create a text of about 300-400 words.\\
2.\hspace{0.5em}The language and format style of the text should be chosen from the given styles.\\
3.\hspace{0.5em}You can insert some LaTeX formulas, choosing from the following formula styles:\\
    \hspace*{1em} - Matrix styles, such as matrix, array, pmatrix, bmatrix, vmatrix, Vmatrix, Bmatrix, cases, rcases, smallmatrix, subarray, etc.\\
    \hspace*{1em} - Multiline formula styles, such as equation\&split, align, gather, alignat, etc.\\
    \hspace*{1em} - Regular mathematical formula styles, such as frac, sum, etc.\\
4.\hspace{0.5em}LaTeX formulas can be divided into inline and display; inline formulas use the '\$...\$' style, and display formulas use the '\$\$...\$\$' style.\\
5.\hspace{0.5em}Both inline and display formulas can be inserted into the generated text.\\
6.\hspace{0.5em}Do not insert any tables in the text.\\
7.\hspace{0.5em}You can choose several of the following MarkDown syntax styles when creating:\\
   \hspace*{1em} - Different levels of headings\\
   \hspace*{1em} - Bold\\
   \hspace*{1em} - Italic\\
   \hspace*{1em} - Bold and italic\\
   \hspace*{1em} - Underline\\
   \hspace*{1em} - Superscript\\
   \hspace*{1em} - Subscript\\
   \hspace*{1em} - Lists\\
     \hspace*{2em} - Unordered lists\\
     \hspace*{2em} - Ordered lists\\
8.\hspace{0.5em}The style and organization of the entire text should be more varied, **do not always add a summary at the end**.\\
9.\hspace{0.5em}The content of this text does not have to be complete; it can be a truncated content.\\
10.\hspace{0.5em}Please only return the generated text content, do not return any other content, such as descriptions before and after.\\
11.\hspace{0.5em}Do not indicate the chosen topic, text language, and format style in the generated text.\\
12.\hspace{0.5em}Please return in English.\\
13.\hspace{0.5em}Random seed is SEED\\

Available text language and format styles:\\
Exam paper, slides, academic paper, book, textbook, magazine, notes, newspaper, financial report\\

Example:\\
\# The Future of Artificial Intelligence\\

**Artificial Intelligence** (AI), as one of the most revolutionary technologies of the 21st century, is rapidly changing our way of life and work patterns. The future of AI is full of infinite possibilities, but it also comes with numerous challenges and ethical issues.\\

\#\# Technological Advancements and Applications\\

The advancements in AI technology are mainly reflected in the following areas:\\

1.\hspace{0.5em}**Deep Learning**: Through multi-layer neural networks, AI can process complex data patterns. For example, Convolutional Neural Networks (CNN) perform excellently in image recognition, while Recurrent Neural Networks (RNN) have wide applications in Natural Language Processing (NLP).\\
2.\hspace{0.5em}**Reinforcement Learning**: AI continuously optimizes the decision-making process through interaction with the environment. A famous example is AlphaGo, which surpassed top human players by playing against itself.\\
3.\hspace{0.5em}**Transfer Learning**: AI can transfer knowledge from one task to another, improving the generalization ability of models.\\

\#\# Application of Mathematical Formulas\\

In AI research, mathematical formulas play a crucial role. For example, the training process of neural networks can be represented by the following formula:\\

\$\$\\
\textbackslash text{Loss} = \textbackslash frac\{1\}\{N\} \textbackslash sum\_\{i=1\}\^\{N\} L(y\_i, \textbackslash hat\{y\}\_i)\\
\$\$\\

where \$L\$ represents the loss function, \$y\_i\$ is the actual value, \$\\hat\{y\}\_i\$ is the predicted value, and \$N\$ is the number of samples.\\

\#\# Ethics and Challenges\\

Although AI brings many conveniences, it also raises ethical and social issues:\\

- **Privacy Issues**: AI may infringe on personal privacy during data processing.\\
- **Employment Impact**: Automation technology may lead to the disappearance of certain jobs, increasing the unemployment rate.\\
- **Decision Transparency**: The black-box nature of AI algorithms makes the decision-making process difficult to explain, potentially leading to unfair outcomes.\\

\$\$\\
\textbackslash text\{CNN\} = \textbackslash left( \textbackslash begin\{array\}\{ccc\}\\
f\_1 \& f\_2 \& f\_3 \textbackslash \textbackslash \\
f\_4 \& f\_5 \& f\_6 \textbackslash \textbackslash \\
f\_7 \& f\_8 \& f\_9 \textbackslash \textbackslash \\
\textbackslash end{array} \textbackslash right)\\
\$\$
\end{promptbox}

\begin{promptbox}[Prompt for Tables]{lightgreen}
Please create a MarkDown text based on the given table and the current random seed, with the following requirements:\\

1.\hspace{0.5em}Create the text based on the content of the table, around 300 words, and insert the table into the generated text.\\
2.\hspace{0.5em}You can choose to use some of the following MarkDown syntax in the creation:\\
   \hspace*{1em} - Different levels of headings\\
   \hspace*{1em} - Bold\\
   \hspace*{1em} - Italic\\
   \hspace*{1em} - Bold Italic\\
   \hspace*{1em} - Underline\\
   \hspace*{1em} - Superscript\\
   \hspace*{1em} - Subscript\\
   \hspace*{1em} - Lists\\
     \hspace*{2em} - Unordered lists\\
     \hspace*{2em} - Ordered lists\\
3.\hspace{0.5em}The style and organization of the whole text should be more varied, **do not always add a summary at the end**\\
4.\hspace{0.5em}The content of this text does not have to be complete, it can be a truncated content\\
5.\hspace{0.5em}Please return only the generated text, do not return any other content, such as descriptions before and after\\
6.\hspace{0.5em}Please return in English\\
7.\hspace{0.5em}Insert the table as it is, do not make any changes to the table, and there should be no line breaks or indentations between the html tags\\
8.\hspace{0.5em}The random seed is: SEED\\

TABLE\\

\end{promptbox}

\begin{promptbox}[Prompt for Multi-column]{lightgreen}
Please choose one of the topics given below and, based on the current random seed, write a MarkDown text with the following requirements:\\

1.\hspace{0.5em}Create content based on the theme "TOPIC", with a word count of around 600-800 words.\\
2.\hspace{0.5em}The language and format style of the text should be chosen from the styles provided below.\\
3.\hspace{0.5em}You may insert some Latex formulas, choosing from the following formula styles:\\
    \hspace*{1em} - Matrix styles, such as matrix, array, pmatrix, bmatrix, vmatrix, Vmatrix, Bmatrix, cases, rcases, smallmatrix, subarray, etc.\\
    \hspace*{1em} - Multi-line equation styles, such as equation\&split, align, gather, alignat, etc.\\
    \hspace*{1em} - Regular mathematical formula styles, such as frac, sum, etc.\\
4. Latex formulas can be divided into inline and display; inline formulas use the '\$...\$' style, while display formulas use the '\$\$...\$\$' style.\\
5. Both inline and display formulas can be inserted into the generated text.\\
6. You may choose several MarkDown syntax options to use during creation:\\
   \hspace*{1em} - Different levels of headings\\
   \hspace*{1em} - Bold\\
   \hspace*{1em} - Italic\\
   \hspace*{1em} - Bold Italic\\
   \hspace*{1em} - Underline\\
   \hspace*{1em} - Superscript\\
   \hspace*{1em} - Subscript\\
   \hspace*{1em} - Lists\\
     \hspace*{2em} - Unordered lists\\
     \hspace*{2em} - Ordered lists\\
7.\hspace{0.5em}The style and organization of the entire text should be more varied; **do not always add concluding remarks at the end**.\\
8.\hspace{0.5em}The content of this text does not have to be complete and can be a truncated piece.\\
9.\hspace{0.5em}Please return only the generated text content, without any additional descriptions before or after.\\
10.\hspace{0.5em}Do not indicate the chosen text language or format style in the returned text.\\
11.\hspace{0.5em}Return the text in English.\\
12.\hspace{0.5em}The random seed is: SEED\\
13.\hspace{0.5em}Divide this text into two paragraphs, with "x----------x" separating the paragraphs.\\

Available text languages and format styles:\\
Exam paper, slides, academic paper, book, textbook, magazine, notes, newspaper, financial report\\

Here is an example:\\

\# The Wonderful World of Mathematics\\

Mathematics, this ancient yet vibrant discipline, runs through the entire development of human civilization. From the geometry of ancient Greece to modern topology, mathematics is not only a tool for solving problems but also a way to explore the essence of the world. **The charm of mathematics** lies in its abstraction and universality—whether it's the laws of nature or phenomena in human society, all can be described and explained using the language of mathematics.\\

In mathematics, matrices are a very important concept. Matrices occupy a central position in linear algebra and are widely used in physics, computer science, economics, and other fields. An \$m \textbackslash times n\$ matrix can be represented as:\\

\$\$\\
A = \textbackslash begin\{pmatrix\}\\
a\_\{11\} \& a\_\{12\} \& \textbackslash cdots \& a\_\{1n\} \textbackslash \textbackslash \\
a\_\{21\} \& a\_\{22\} \& \textbackslash cdots \& a\_\{2n\} \textbackslash \textbackslash \\
\textbackslash vdots \& \textbackslash vdots \& \textbackslash ddots \& \textbackslash vdots \textbackslash \textbackslash \\
a\_\{m1\} \& a\_\{m2\} \& \textbackslash cdots \& a\_\{mn\}
\textbackslash end\{pmatrix\}\\
\$\$\\

Matrix operations include addition, subtraction, multiplication, and transposition. In particular, matrix multiplication is not just a simple element-by-element multiplication but involves the inner product of rows and columns. Suppose there are two matrices \$A\$ and \$B\$, their product \$C\$ can be expressed as:\\

\$\$\\
C = A \textbackslash cdot B\\
\$\$\\

Here, the elements \$c\_\{ij\}\$ of \$C\$ are obtained by the inner product of the \$i\$-th row of \$A\$ and the \$j\$-th column of \$B\$:\\

\$\$\\
c\_\{ij\} = \textbackslash sum\_\{k=1\}\^\{n\} a\_\{ik\} b\_\{kj\}\\
\$\$\\

x----------x\\

Besides matrices, there are many other important concepts and tools in mathematics. For example, calculus is a mathematical tool for studying change and has wide applications in physics, engineering, and economics. The fundamental idea of calculus is to study the rate of change and accumulation of functions through the concept of limits. The derivative \$f'(x)\$ of a function \$f(x)\$ represents the instantaneous rate of change of the function at point \$x\$, which can be expressed in limit form as:\\

\$\$\\
f'(x) = \textbackslash lim\_\{\textbackslash Delta x \textbackslash to 0\} \textbackslash frac\{f(x + \textbackslash Delta x) - f(x)\}\{\textbackslash Delta x\}\\
\$\$\\

Integration is the inverse operation of differentiation, used to calculate the accumulation of a function over an interval. The definite integral of a function \$f(x)\$ over the interval \$[a, b]\$ is expressed as:\\

\$\$\\
\textbackslash int\_\{a\}\^\{b\} f(x) \textbackslash , dx\\
\$\$\\

The beauty of mathematics lies not only in its rigor and logic but also in its simplicity and elegance. Many mathematical theorems and formulas are renowned for their concise forms and profound implications. For instance, Euler's formula \$e\^\{i\textbackslash pi\} + 1 = 0\$ ingeniously connects five seemingly unrelated mathematical constants—the base of the natural logarithm \$e\$, the imaginary unit \$i\$, the circle constant \$\textbackslash pi\$, 1, and 0—showcasing the inherent harmony of mathematics.\\

\end{promptbox}


\subsection{Examples of Unified Formats}

\label{sec:exam_unified}

\begin{promptbox}[Plain Text]{lightgreen}
{\color{red}
\# The Role of Urban Green Spaces in Residents' Mental Health\\

Urban green spaces, such as parks, gardens, and tree-lined streets, play a crucial role in the mental health of city dwellers. These areas are not just aesthetically pleasing; they are vital for psychological well-being, offering a respite from the urban environment's constant hustle and bustle. This excerpt explores the various ways in which urban green spaces contribute to mental health and the mechanisms behind these benefits.\\

\#\# The Therapeutic Effects of Nature\\

\#\#\# Stress Reduction\\
One of the most significant benefits of urban green spaces is their ability to reduce stress. **Research has shown** that spending time in natural environments can lower cortisol levels, a hormone associated with stress. The serene and calming atmosphere of parks and gardens provides a stark contrast to the often chaotic and noisy urban settings. This shift in environment can help individuals relax and regain a sense of peace.\\

\#\#\# Mood Enhancement\\
Urban green spaces also have a positive impact on mood. *Interacting with nature* can boost feelings of happiness and well-being. The presence of greenery and natural elements can help alleviate symptoms of depression and anxiety. Activities such as walking, jogging, or simply sitting in a park can enhance one's mood and provide a sense of tranquility.\\

\#\# Physical Activity and Mental Health\\

\#\#\# Encouraging Physical Activity\\
Urban green spaces often serve as venues for physical activities, which are essential for maintaining good mental health. Parks and recreational areas provide opportunities for residents to engage in exercise, such as walking, running, cycling, }{\color{red}and team sports. **Regular physical activity** has been linked to reduced rates of depression and anxiety, as well as improved cognitive function.\\

\#\#\# Social Interaction\\
These spaces also facilitate social interaction, which is crucial for mental health. Community gardens, playgrounds, and public parks bring people together, fostering a sense of community and belonging. Social connections can provide emotional support and reduce feelings of isolation, which are common in urban environments.\\

\#\# Cognitive Benefits\\

\#\#\# Improved Attention and Focus\\
Urban green spaces can also enhance cognitive function. The natural environment has a restorative effect on mental fatigue, helping individuals to concentrate better and perform tasks more efficiently. This is particularly important in urban areas where constant exposure to stimuli can lead to cognitive overload.\\

\#\#\# Creativity and Problem-Solving\\
Spending time in nature has been shown to boost creativity and problem-solving skills. The change of scenery and the presence of natural elements can inspire new ideas and perspectives. For students and professionals, urban green spaces can serve as a source of inspiration and a place to recharge.\\

\#\# Conclusion\\

While the benefits of urban green spaces are well-documented, it is essential to recognize that these areas are not a one-size-fits-all solution. The design and accessibility of green spaces can significantly influence their effectiveness in promoting mental health. **Urban planners and policymakers** must prioritize the creation and maintenance of green spaces to ensure that all residents have access to these vital resources. By doing so, cities can become more livable and supportive environments for their inhabitants.\\
}
\end{promptbox}

\begin{promptbox}[Tables]{lightgreen}
\# Analysis of Sleep Apnea Metrics in Different Age Groups\\

Sleep apnea is a common disorder that affects both young and middle-aged individuals as well as the elderly. The table below provides a detailed comparison of various sleep apnea metrics between these two age groups, highlighting significant differences in certain parameters.\\

{\color{red}
<table><thead><tr><td></td><td><b>Young and middle age (<i>n</i>= 568, m ± SD)</b></td><td><b>Elderly (<i>n</i>= 129, m ± SD)</b></td><td><b><i>P</i>value</b></td></tr></thead><tbody><tr><td>AHI (events per hour)</td><td>35.14 ± 13.26</td><td>33.14 ± 12.09</td><td>NS</td></tr><tr><td>AHI-REM (events per hour)</td><td>30.72 ± 9.32</td><td>28.16 ± 7.89</td><td>NS</td></tr><tr><td>AHI-NREM (events per hour)</td><td>33.82 ± 12.14</td><td>31.73 ± 11.09</td><td>NS</td></tr><tr><td>Mean apnea/hypopnea duration (sec)</td><td>20.12 ± 6.12</td><td>22.37 ± 6.09</td><td>\&lt;0.001*</td></tr><tr><td>Hypoxemia duration in TST (min)</td><td>51.1 ± 19.12</td><td>73.50 ± 19.07</td><td>0.004*</td></tr><tr><td>Hypoxemia duration in REM sleep (min)</td><td>8.88 ± 2.46</td><td>7.98 ± 2.68</td><td>NS</td></tr><tr><td>Hypoxemia duration in NREM sleep (min)</td><td>28.08 ± 5.43</td><td>42.43 ± 6.98</td><td>0.021*</td></tr><tr><td>SatO<sub>2</sub>-REM</td><td>91.89 ± 7.12</td><td>89.91 ± 7.02</td><td>0.013*</td></tr><tr><td>SatO<sub>2</sub>-NREM</td><td>92.64 ± 5.21</td><td>90.32 ± 5.11</td><td>0.027*</td></tr></tbody></table>\\}
**Table 1: Comparison of Sleep Apnea Metrics Between Young and Middle-Age vs. Elderly Groups**\\
\end{promptbox}

\begin{promptbox}[Mathematical Formulas]{lightgreen}

\#\#\# Geometric Shapes\\

The face can be broken down into basic geometric shapes, which can be mathematically defined:\\

- **Eyes**: Circles or ellipses, with a radius {\color{red}\$r\$} and a center at {\color{red}\$(x, y)\$}.\\
- **Nose**: A triangle or a small circle, with a base {\color{red}\$b\$} and height {\color{red}\$h\$}.\\
- **Mouth**: A parabolic curve, defined by the equation {\color{red}\$y = ax\^2 + bx + c\$}.\\

{\color{red}
\$\$\\
\textbackslash text\{Eyes\} = \textbackslash left( \textbackslash begin\{array\}\{c\}\\
(x\_1, y\_1) \textbackslash \textbackslash \\
(x\_2, y\_2) \textbackslash \textbackslash \\
\textbackslash end\{array\} \textbackslash right)\\
\$\$\\
}
{\color{red}
\$\$\\
\textbackslash text\{Nose\} = \textbackslash left( \textbackslash begin\{array\}\{c\}\\
(x\_3, y\_3) \textbackslash \textbackslash \\
\textbackslash end\{array\} \textbackslash right)\\
\$\$\\

\$\$\\
\textbackslash text\{Mouth\} = y = ax\^2 + bx + c\\
\$\$\\
}








\end{promptbox}

\subsection{Data Samples Constructed during the Iterative Self-improvement Stage}

\input{figures/appendix_figure}

%% file: figures/figure_5.tex
\begin{figure}[!htbp]
\centering
\includegraphics[width=\linewidth,scale=1.00]{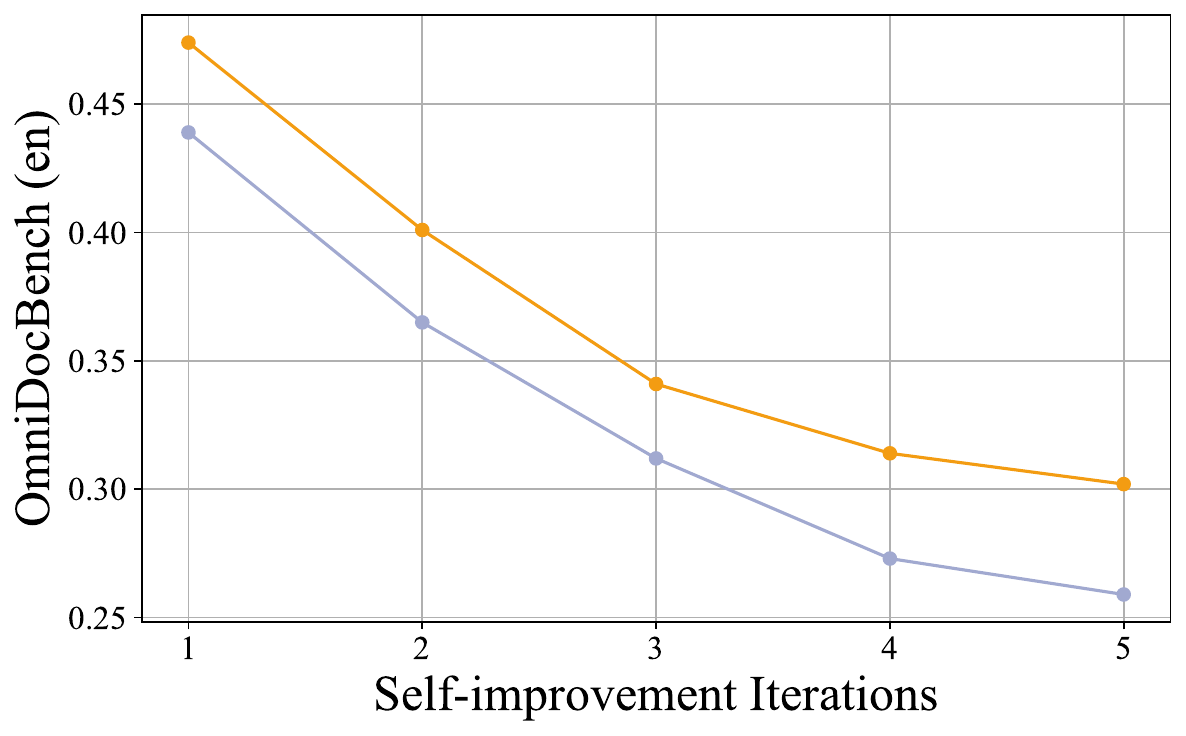}
\caption{\textbf{Loading weights from a previous model can degrade performance.} The orange line and the purple line represent the performance of models initialized from the previous version and the pre-trained model, respectively.}
\label{fig:fig5}
\end{figure}

%% file: figures/figure_6.tex
\begin{figure}[!htbp]
\centering
\includegraphics[width=\linewidth,scale=1.00]{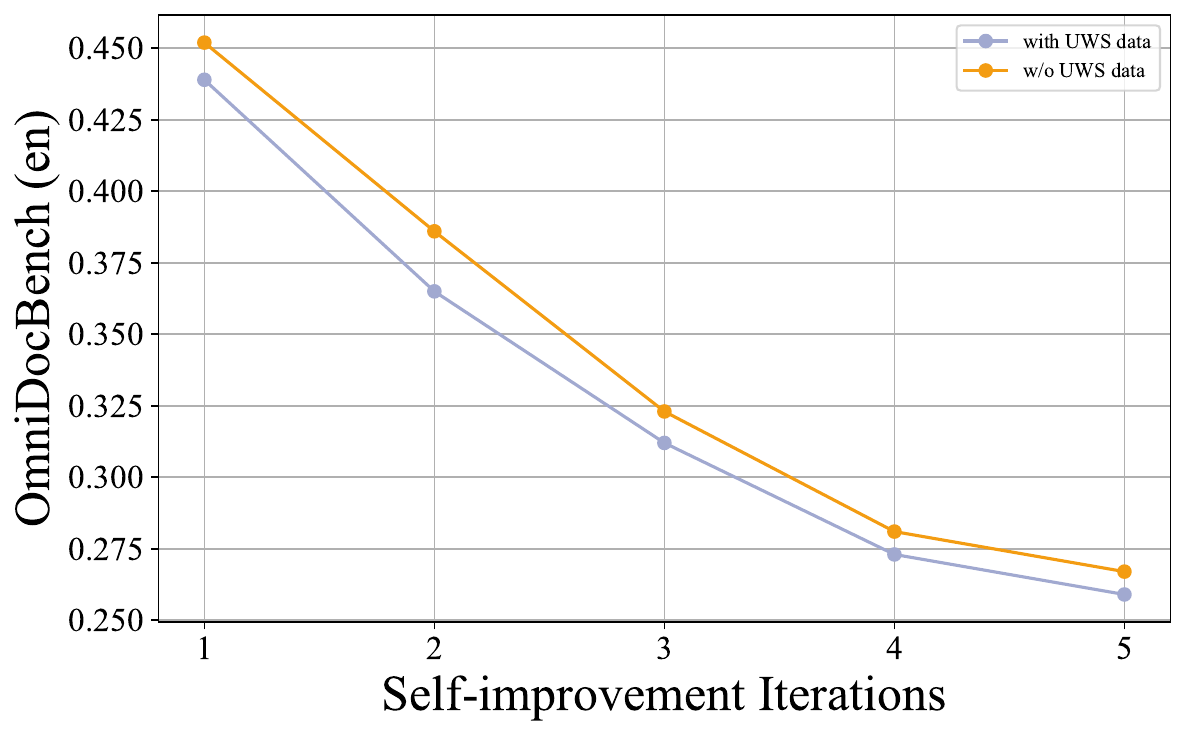}
\caption{\textbf{Including data generated in the UWS into the ISS will benefits the performance of the model.} We include data generated from UWS by default during the iterative self-improvement stage.}
\label{fig:fig5}
\end{figure}

%% file: figures/figure_7.tex
\begin{figure*}[!htbp]
\centering
\includegraphics[width=\linewidth,scale=1.00]{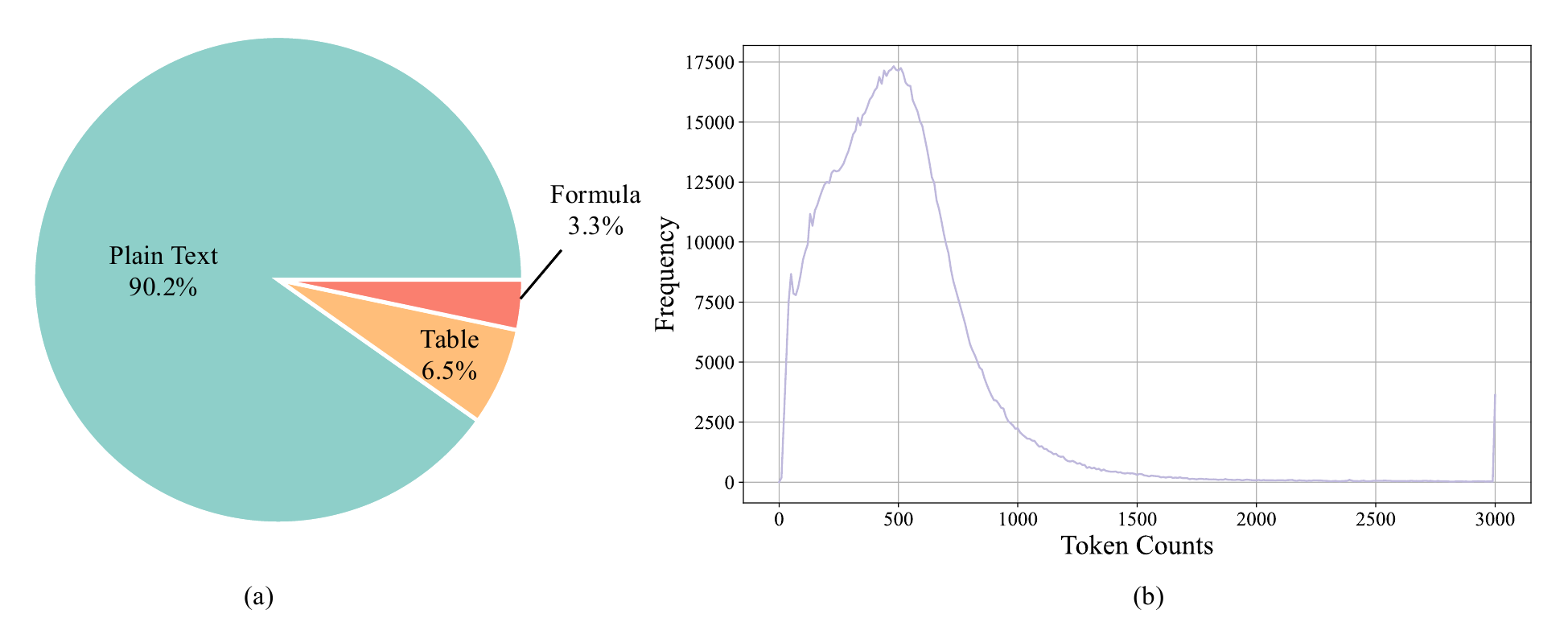}
\caption{\textbf{Distribution of data in the final iteration of the Self-improvement Stage.} (a) shows the proportion of samples containing only plain text, formulas, and tables. (b) presents the distribution of sample counts with respect to different token lengths.}
\label{fig:fig7}
\end{figure*}

%% file: tables/table_6.tex
\begin{table}[ht]
\centering
\tabcolsep 1.5pt
\scalebox{0.85}{
\begin{tabular}{lccccc}
\rowcolor{gray!30}
Data&Text$\downarrow$&Table$\downarrow$&Formula$\downarrow$& Order$\downarrow$&Overall$\downarrow$\\
\hline
Distill & 0.189 & 0.319 & 0.507 & 0.195 & 0.302 \\
POINTS-Reader & 0.176 & 0.383 & 0.335 & 0.144 & 0.259 \\
\end{tabular}
}
\vspace{-0.5em}
\caption{\textbf{Comparison with direct distillation from Qwen2.5-VL-72B}. } 
\label{tab:tab6}
\end{table}

%% file: tables/table_7.tex
\begin{table}[ht]
\centering
\tabcolsep 2pt
\scalebox{0.77}{
\begin{tabular}{lllccccc}
\rowcolor{gray!30}
Plain&Table&Formula&Text$\downarrow$&Table$\downarrow$&Formula$\downarrow$& Order$\downarrow$&Overall$\downarrow$\\
1.00 & 1.00 & 1.00 & 0.176 & 0.383 & 0.335 & 0.144 & 0.259 \\
0.50 & 2.00 & 4.00 & 0.285 & 0.381 & 0.286 & 0.266 & 0.305 \\
0.25 & 4.00 & 8.00 & 0.294 & 0.380 & 0.286 & 0.276 & 0.309 \\
\end{tabular}
}
\vspace{-0.5em}
\caption{\textbf{Ablation about the sampling ratio for different types of samples.} Plain: plain text.} 
\label{tab:tab7}
\end{table}

%% file: figures/figure_8.tex
\begin{figure*}[!h]
\centering
\includegraphics[width=\linewidth,scale=1.00]{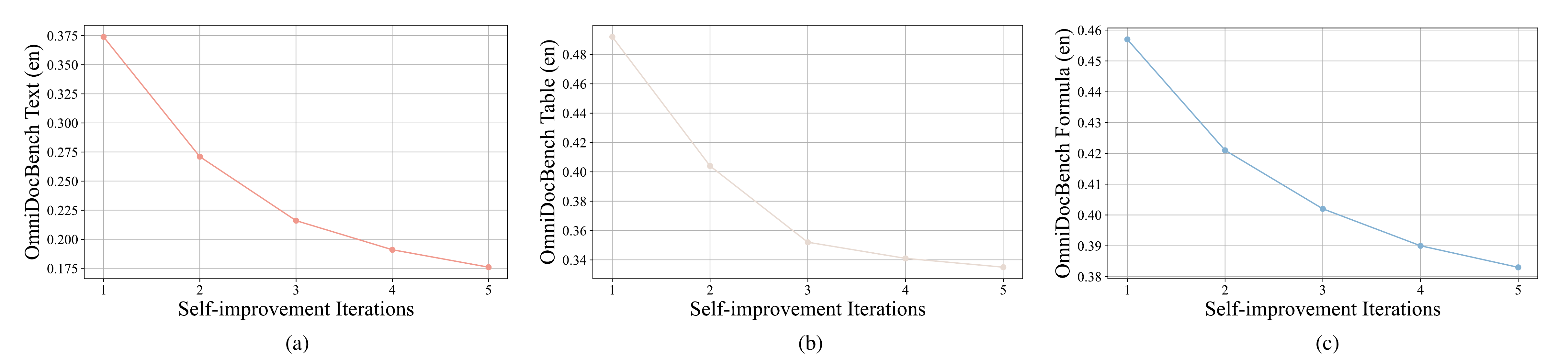}
\caption{\textbf{Steady improvement of performance on plain text, table and formula during the self-improvement stage.}}
\label{fig:fig8}
\end{figure*}

%% file: figures/figure_12.tex
\begin{figure*}[!h]
\centering
\includegraphics[width=\linewidth,scale=1.00]{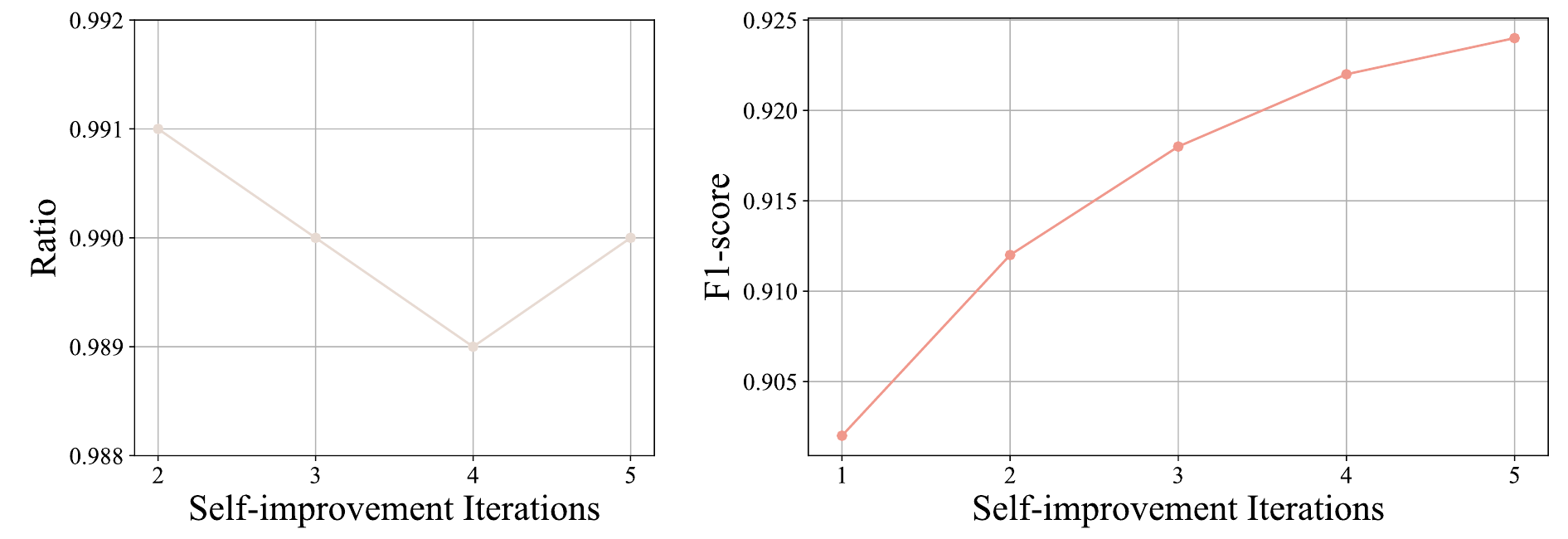}
\caption{The retained ratio of data from previous iteration (left) and the F1-score computed on }
\label{fig:fig12}
\end{figure*}

%% file: tables/table_8.tex
\begin{table}[ht]
\centering
\tabcolsep 2pt
\scalebox{0.77}{
\begin{tabular}{lcccc}
\rowcolor{gray!30}
Method & \#Samples & Table & Distill & Lan \\
KOSMOS-2.5\citep{lv2023kosmos} & 357.4M & MD & Yes & EN \\
olmOCR\citep{poznanski2025olmocr} & 260,000 & MD & Yes & EN \\
POINTS-Reader & 1.1M & HTML & No & EN\\
\end{tabular}
}
\vspace{-0.5em}
\caption{\textbf{Comparison with datasets used in other works.} Table: table format. Distill: whether use distilled data from other models. MD: Markdown.} 
\label{tab:tab8}
\end{table}

%% file: figures/figure_9.tex
\begin{figure*}[!h]
\centering
\includegraphics[width=\linewidth,scale=1.00]{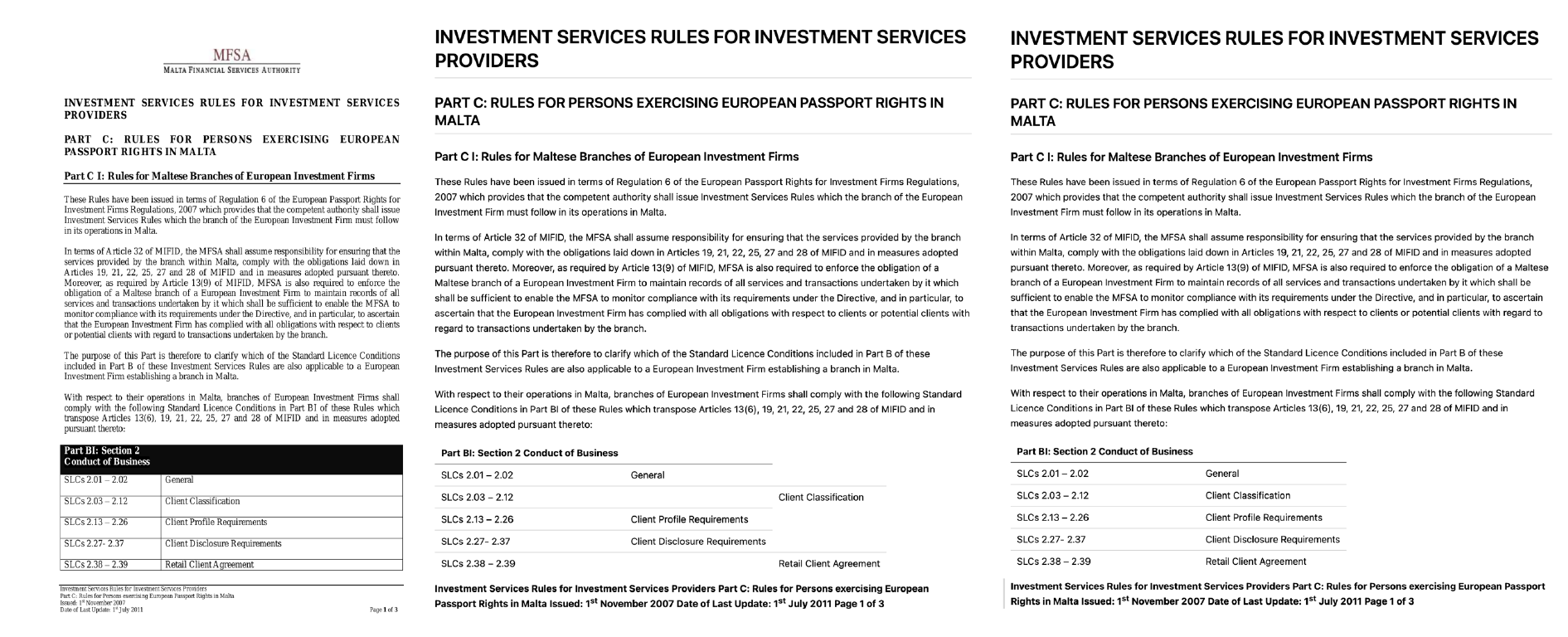}
\caption{\textbf{Case study of samples evolved during the self-improvement stage.} The first figure shows the original document, the second figure presents the annotation generated by the model in the first iteration, and the last figure displays the annotation produced by the model in the final iteration.}
\label{fig:fig9}
\end{figure*}

%% file: figures/figure_10.tex
\begin{figure*}[!h]
\centering
\includegraphics[width=\linewidth,scale=1.00]{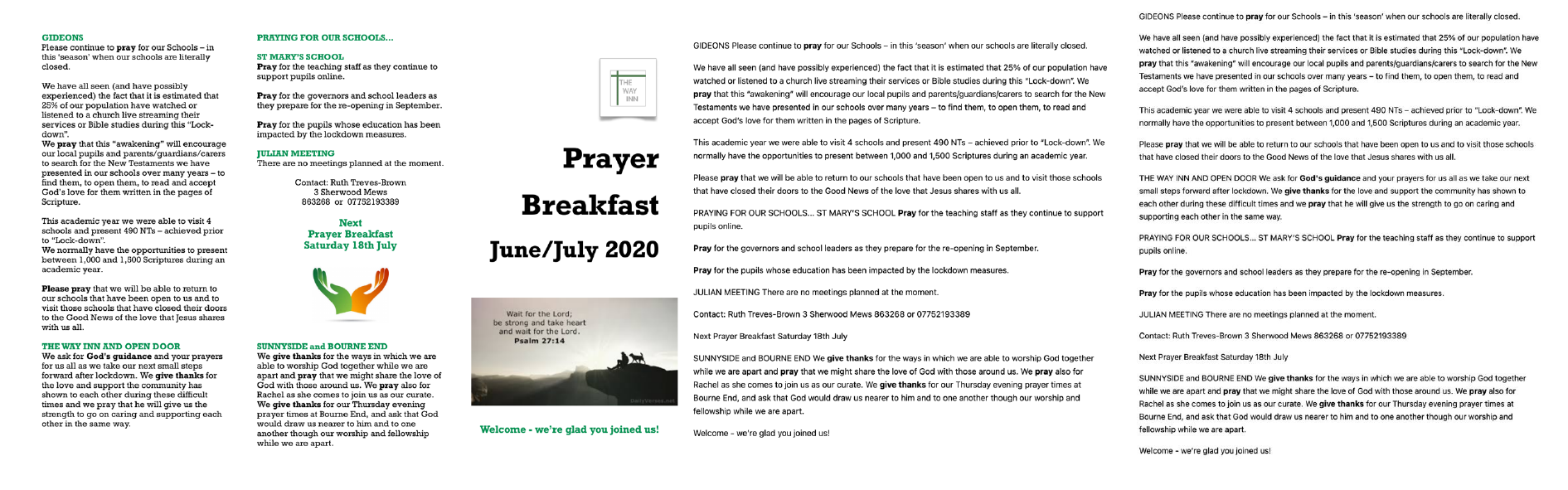}
\caption{\textbf{Case study of samples evolved during the self-improvement stage.} The first figure shows the original document, the second figure presents the annotation generated by the model in the first iteration, and the last figure displays the annotation produced by the model in the final iteration.}
\label{fig:fig10}
\end{figure*}

%% file: figures/figure_11.tex
\begin{figure*}[!h]
\centering
\includegraphics[width=\linewidth,scale=1.00]{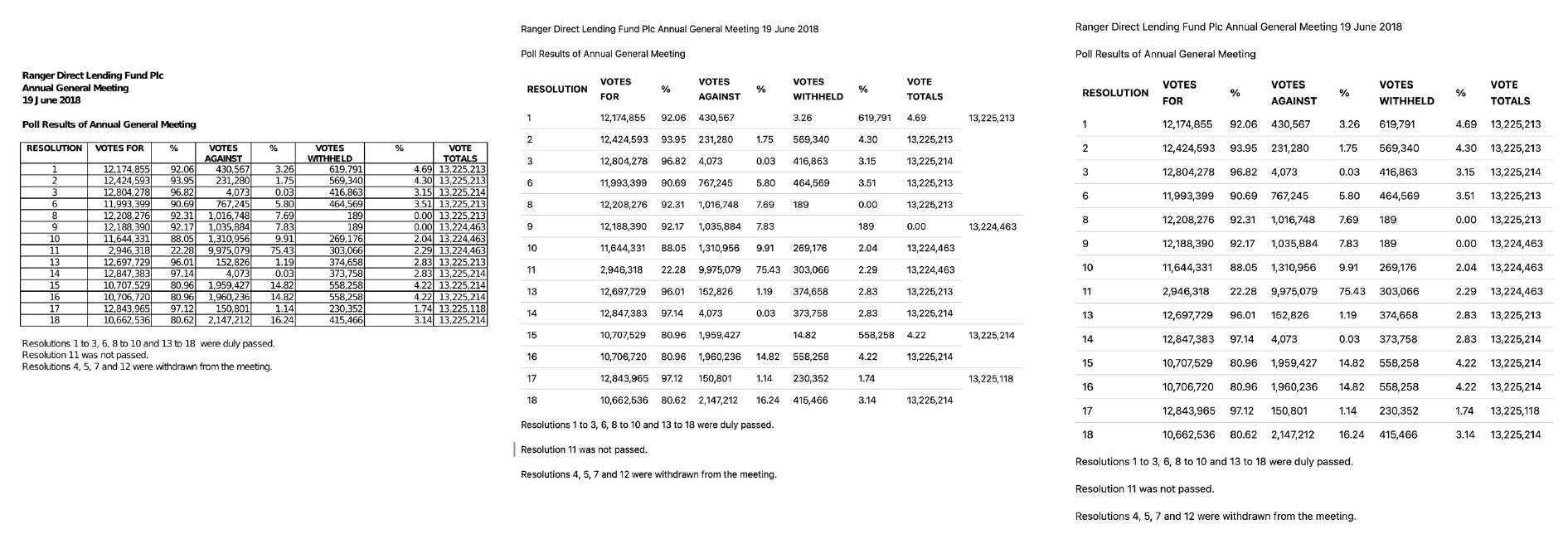}
\caption{\textbf{Case study of samples evolved during the self-improvement stage.} The first figure shows the original document, the second figure presents the annotation generated by the model in the first iteration, and the last figure displays the annotation produced by the model in the final iteration.}
\label{fig:fig10}
\end{figure*}

%% file: figures/appendix_figure.tex
\begin{figure*}[!htbp]
\centering
\includegraphics[width=0.9\linewidth,scale=1.00]{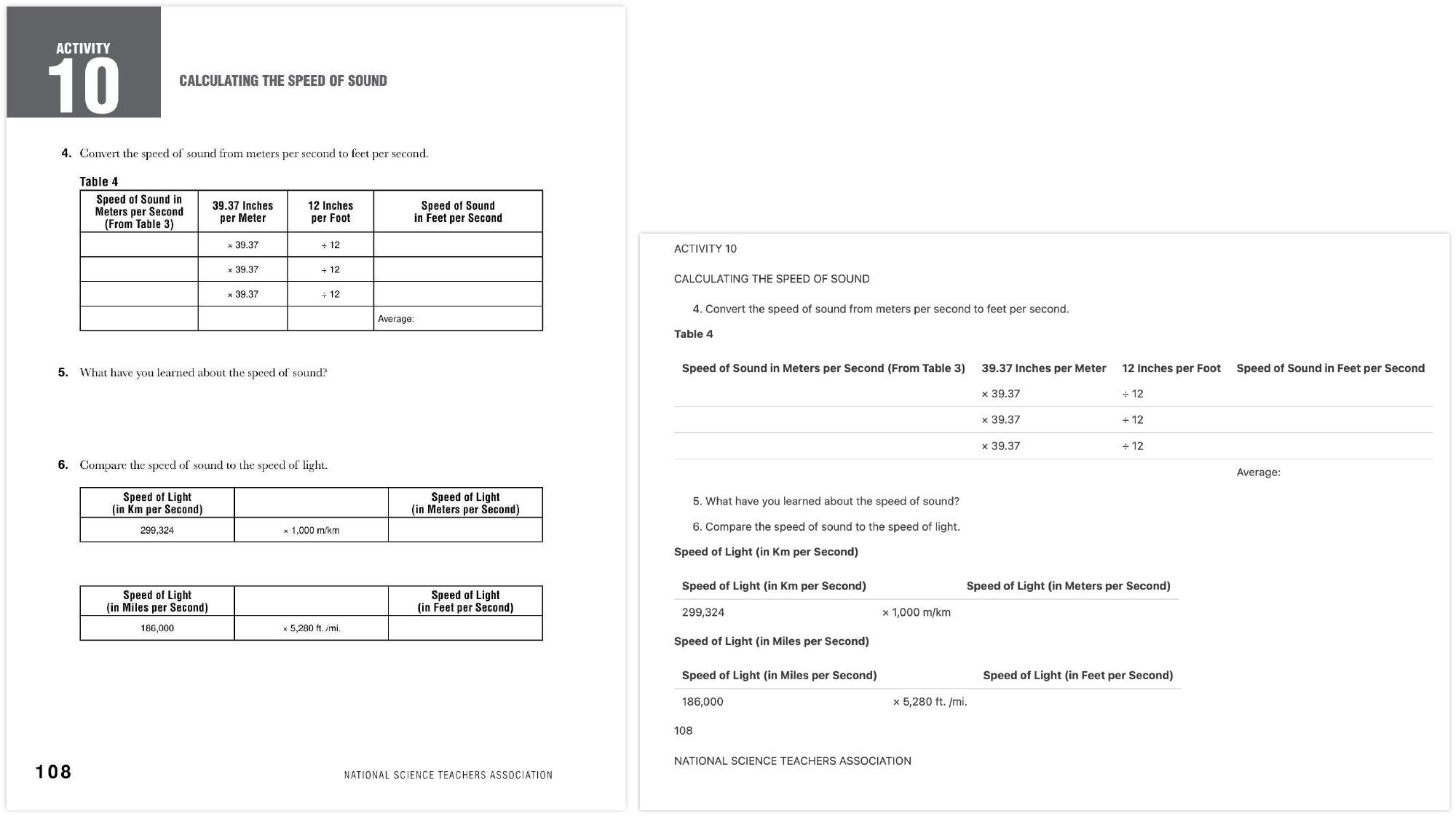}
\caption{The left image displays a document from DocMatix, while the right image presents the corresponding text, rendered as an image and automatically generated by our model}
\label{fig:af1}
\end{figure*}

\begin{figure*}[!htbp]
\centering
\includegraphics[width=0.9\linewidth,scale=1.00]{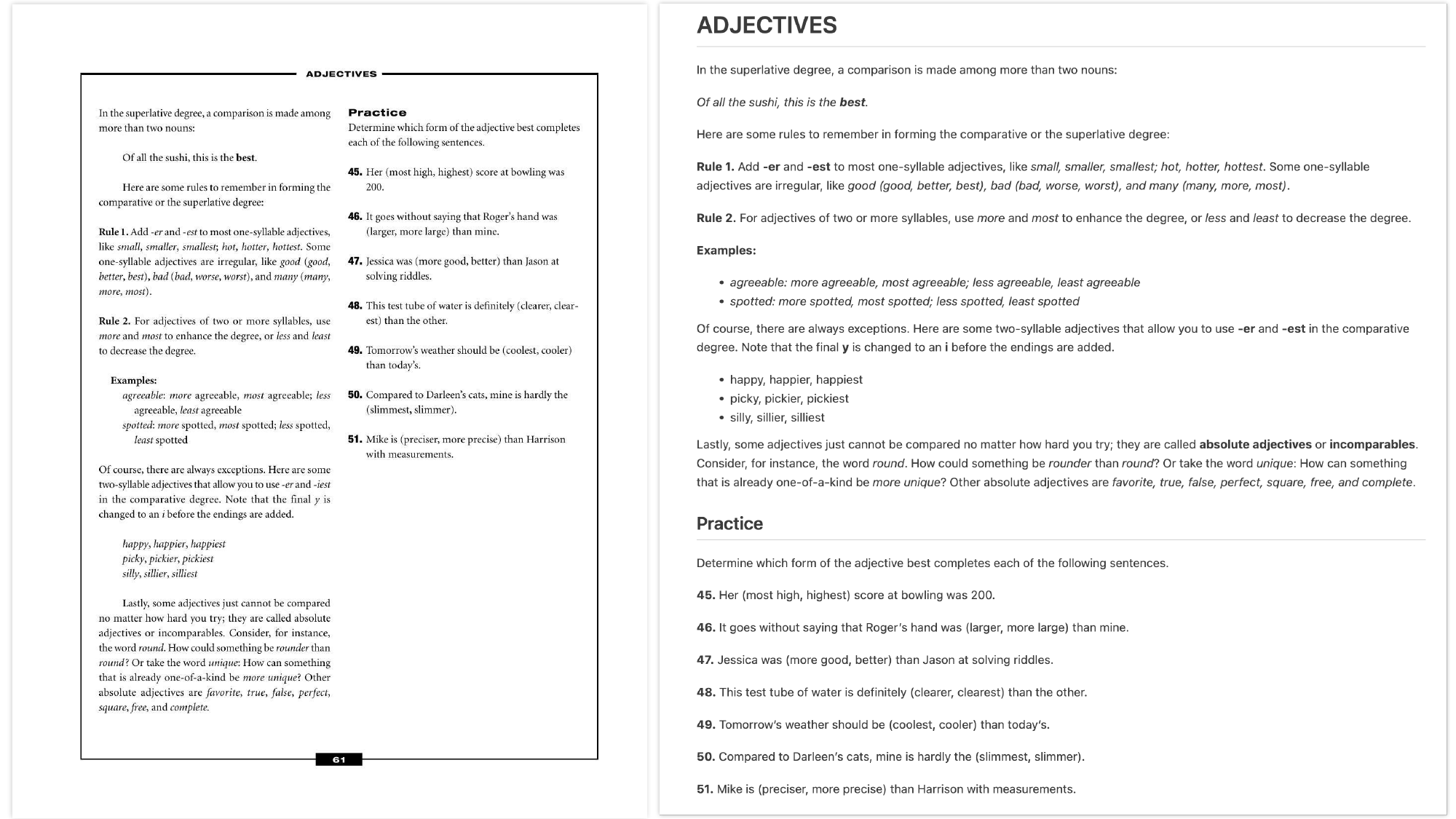}
\caption{The left image displays a document from DocMatix, while the right image presents the corresponding text, rendered as an image and automatically generated by our model}
\label{fig:af2}
\end{figure*}

\begin{figure}[!htbp]
\centering
\includegraphics[width=\linewidth,scale=1.00]{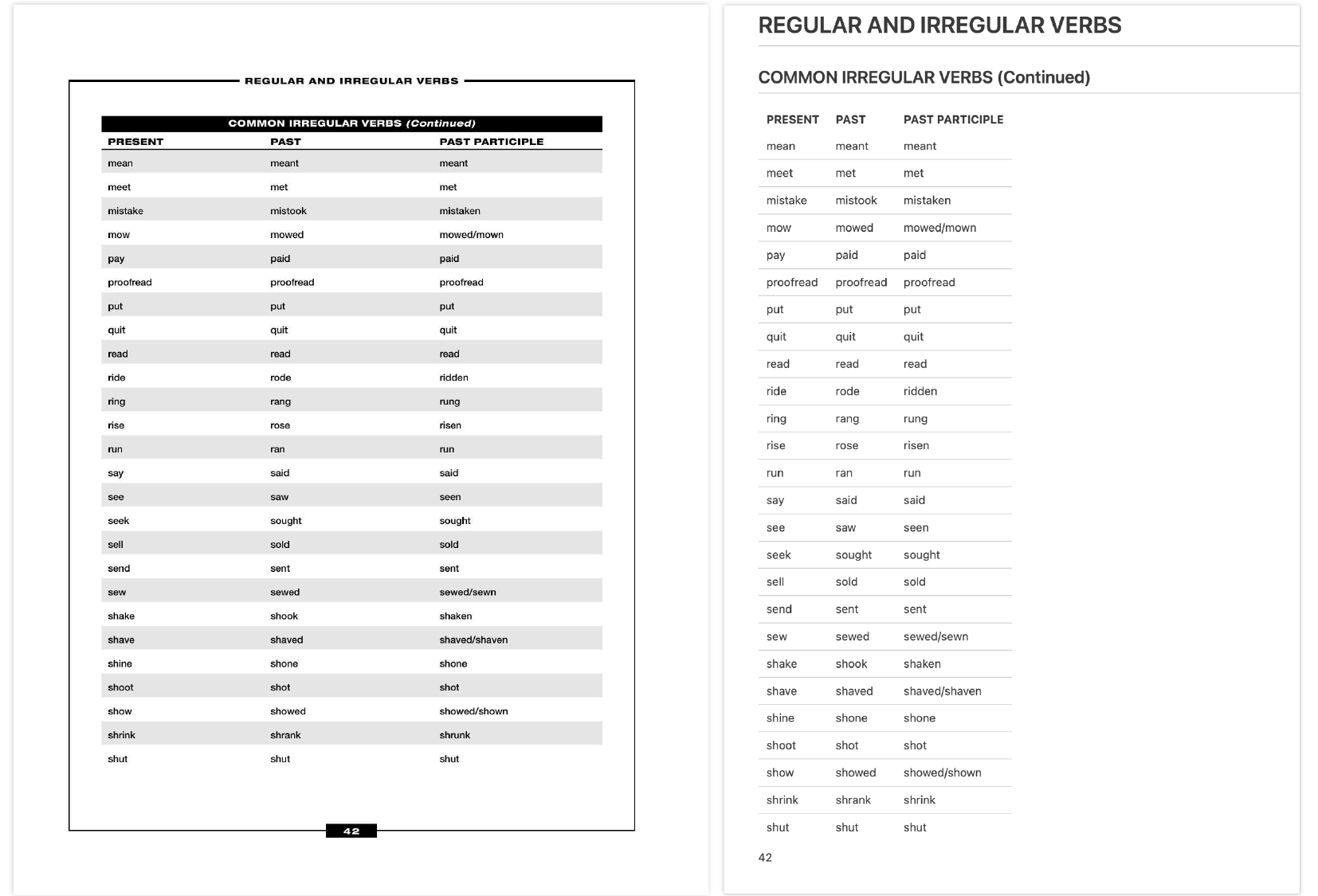}
\caption{The left image displays a document from DocMatix, while the right image presents the corresponding text, rendered as an image and automatically generated by our model}
\label{fig:af3}
\end{figure}

\begin{figure}[!htbp]
\centering
\includegraphics[width=\linewidth,scale=1.00]{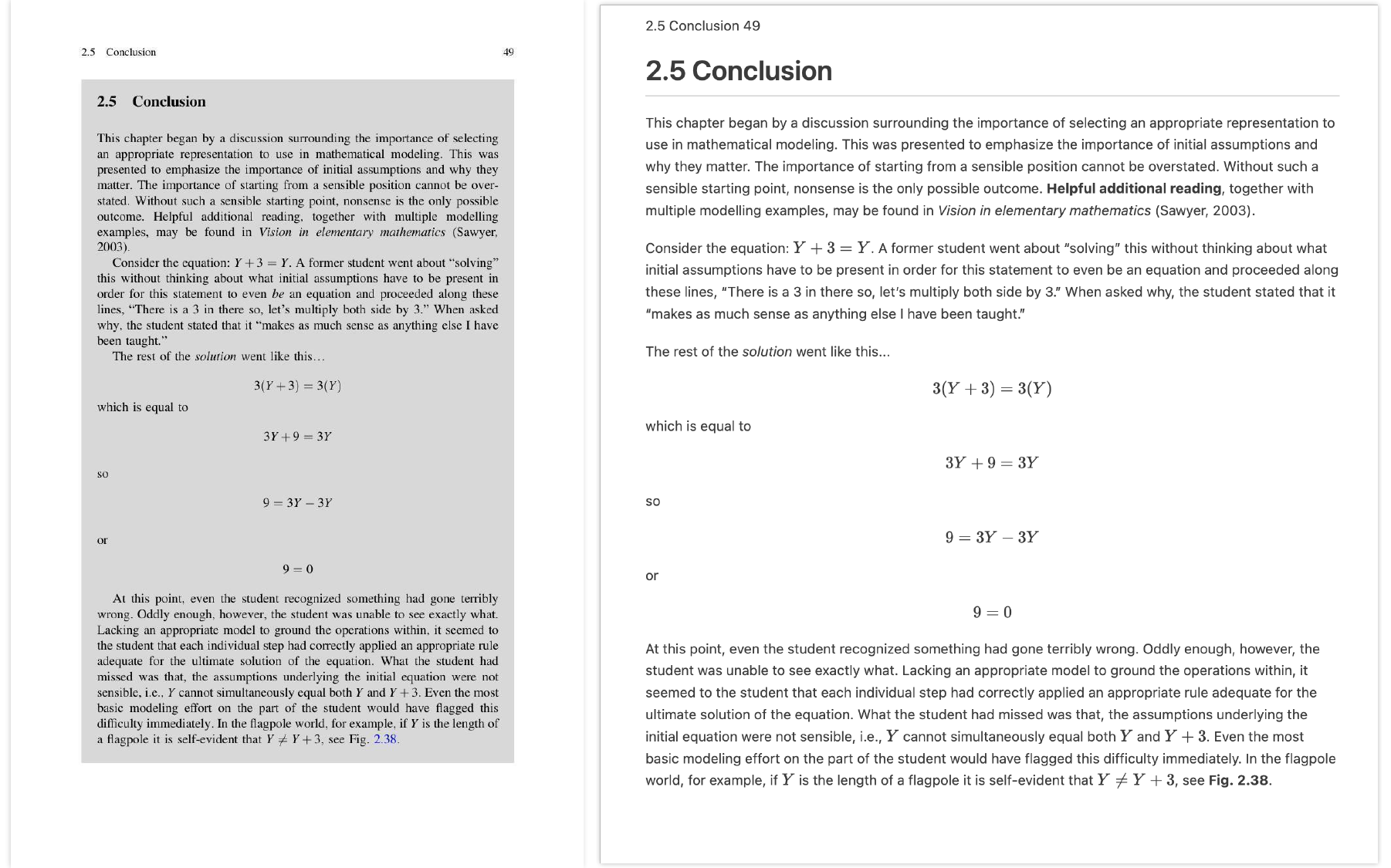}
\caption{The left image displays a document from DocMatix, while the right image presents the corresponding text, rendered as an image and automatically generated by our model}
\label{fig:af4}
\end{figure}

\begin{figure}[!htbp]
\centering
\includegraphics[width=\linewidth,scale=1.00]{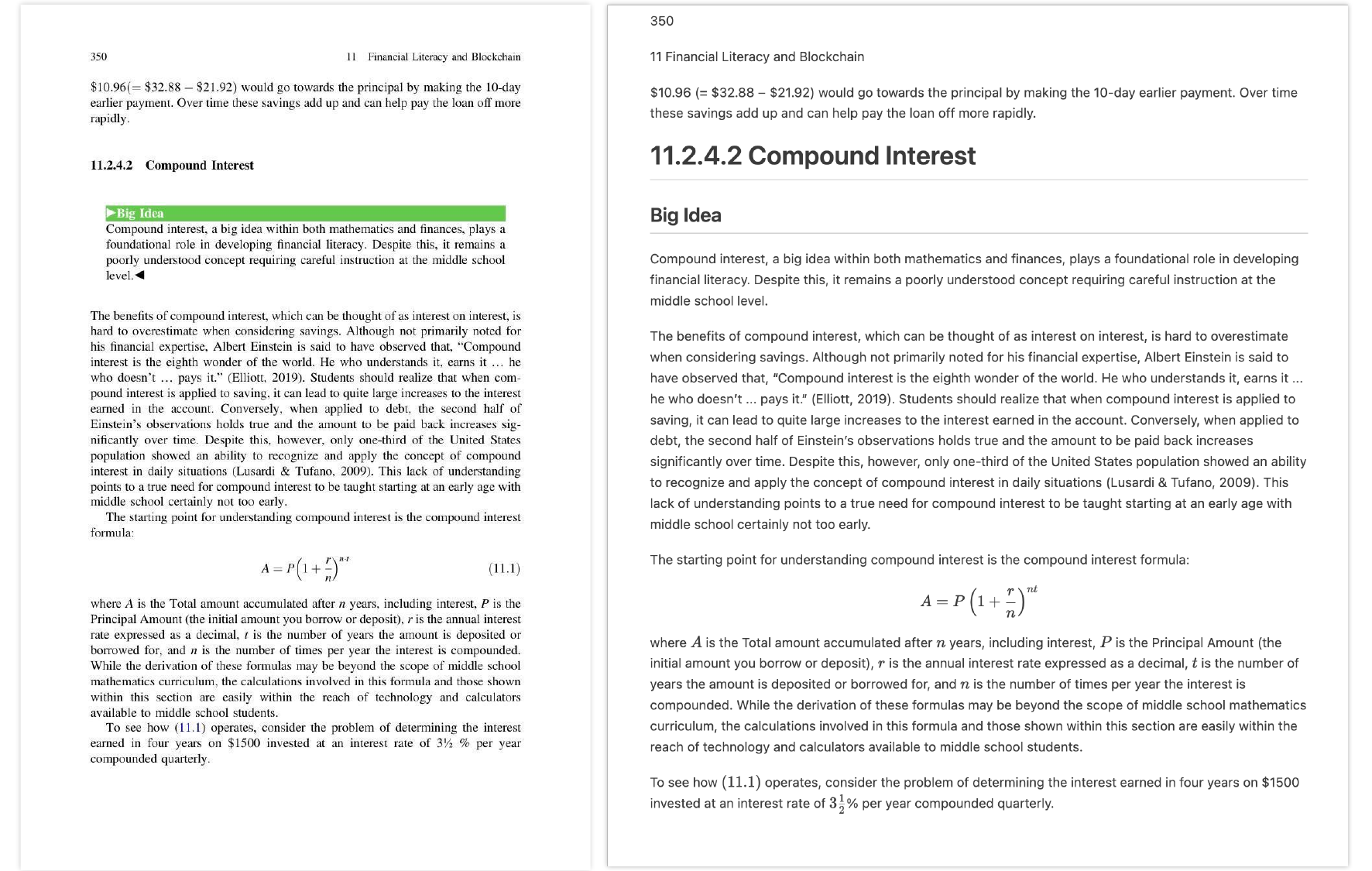}
\caption{The left image displays a document from DocMatix, while the right image presents the corresponding text, rendered as an image and automatically generated by our model}
\label{fig:af5}
\end{figure}

\begin{figure}[!htbp]
\centering
\includegraphics[width=\linewidth,scale=1.00]{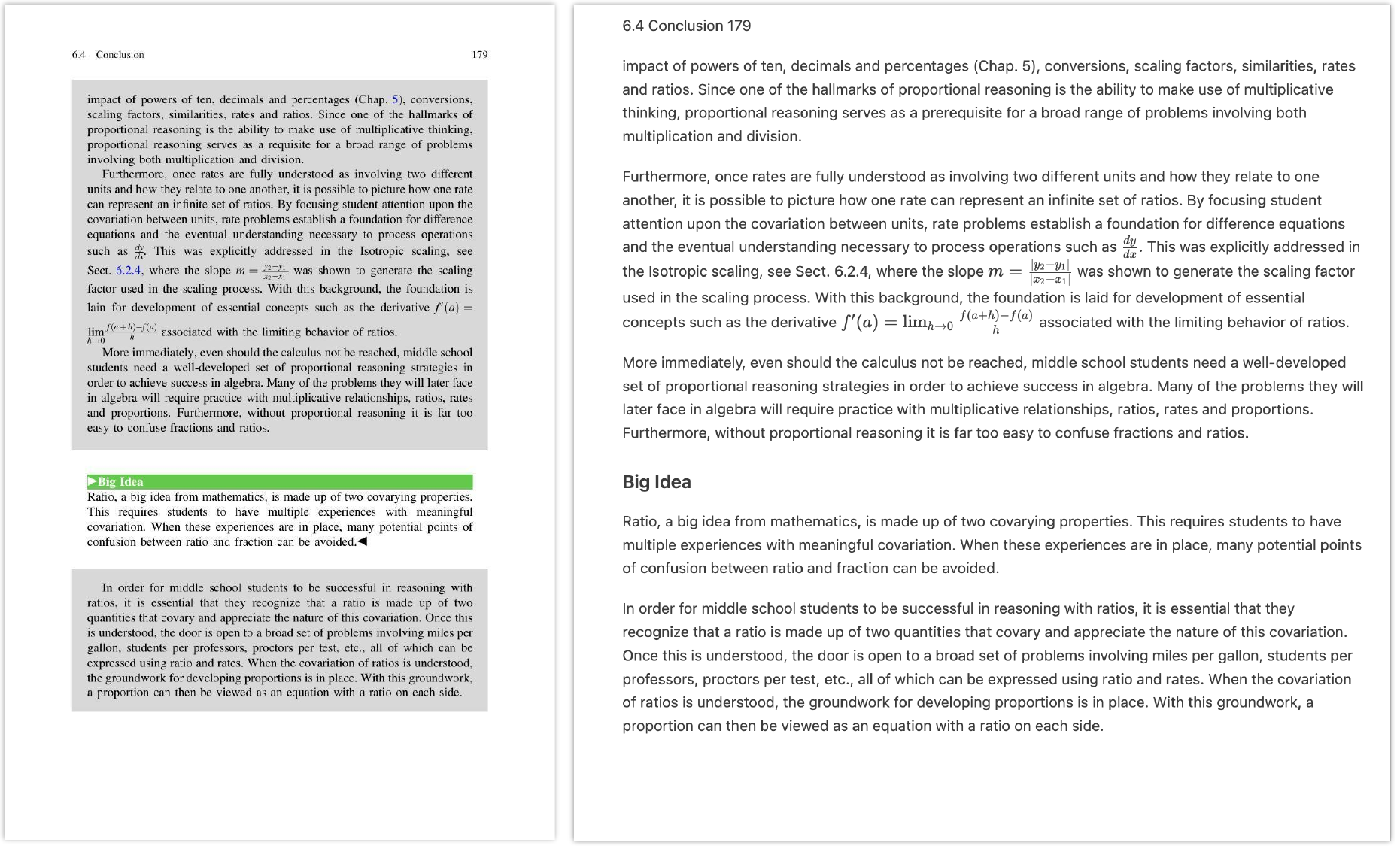}
\caption{The left image displays a document from DocMatix, while the right image presents the corresponding text, rendered as an image and automatically generated by our model}
\label{fig:af6}
\end{figure}

\begin{figure}[!htbp]
\centering
\includegraphics[width=\linewidth,scale=1.00]{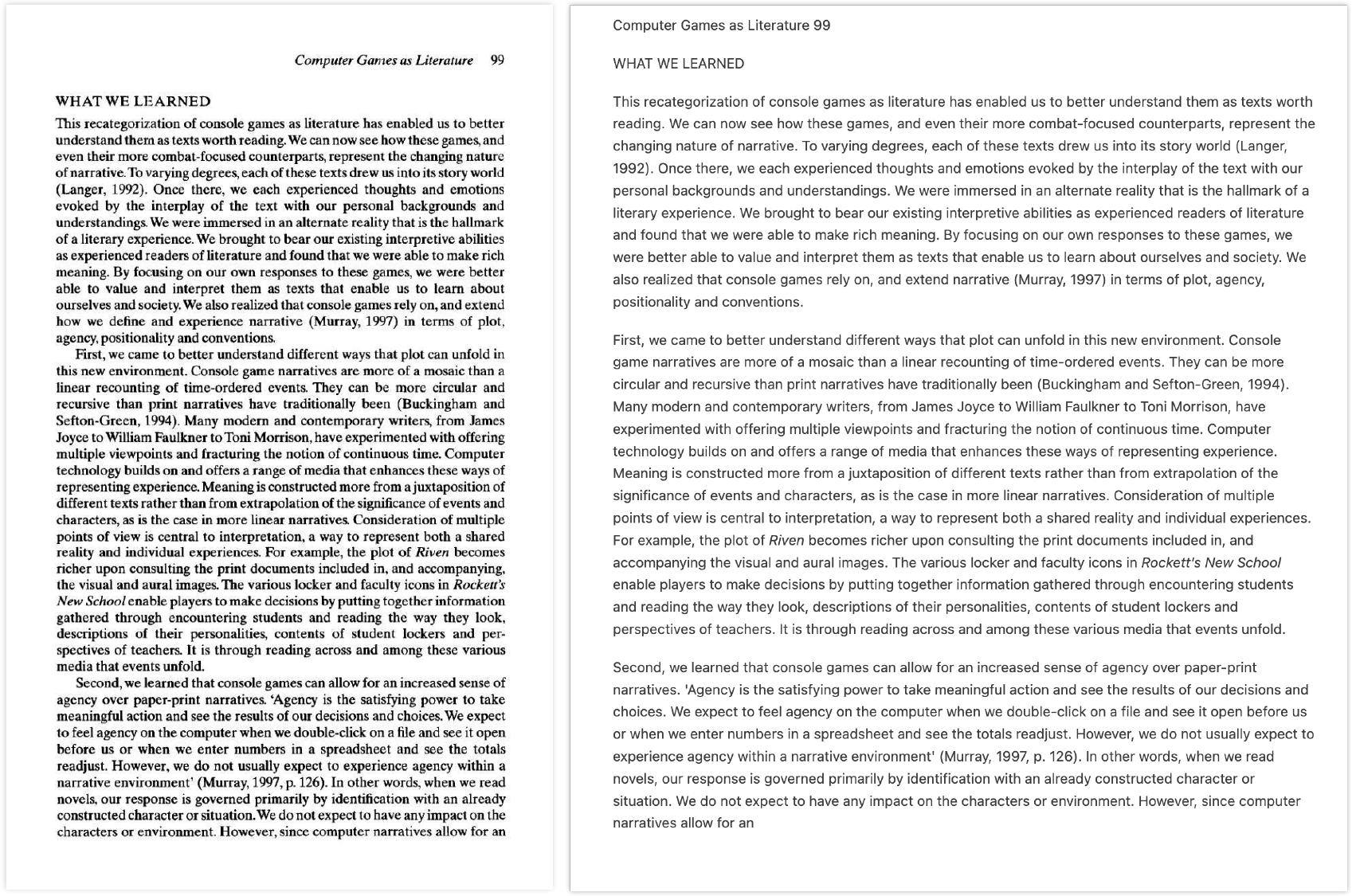}
\caption{The left image displays a document from DocMatix, while the right image presents the corresponding text, rendered as an image and automatically generated by our model}
\label{fig:af7}
\end{figure}

\begin{figure}[!htbp]
\centering
\includegraphics[width=\linewidth,scale=1.00]{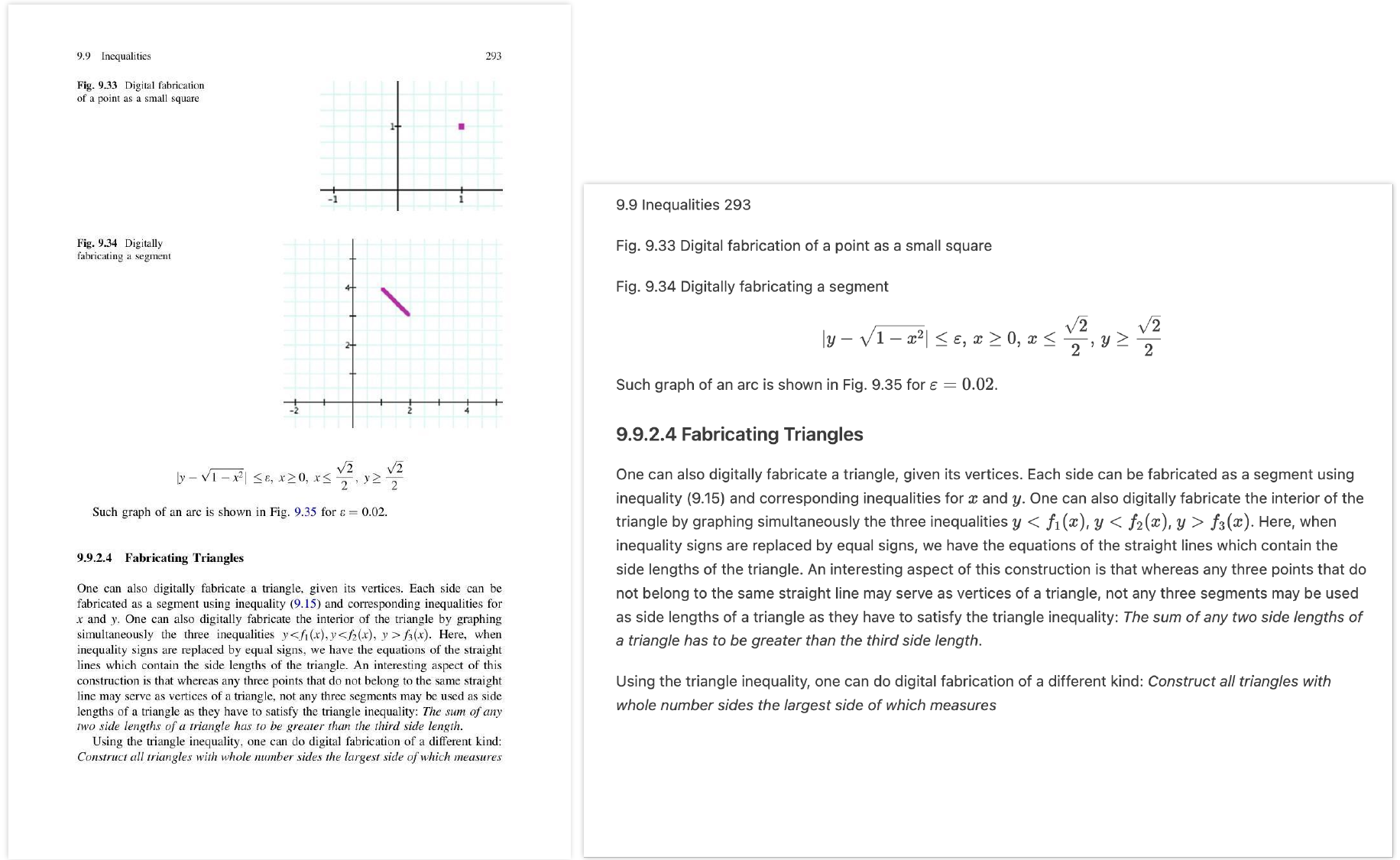}
\caption{The left image displays a document from DocMatix, while the right image presents the corresponding text, rendered as an image and automatically generated by our model}
\label{fig:af8}
\end{figure}

\begin{figure}[!htbp]
\centering
\includegraphics[width=\linewidth,scale=1.00]{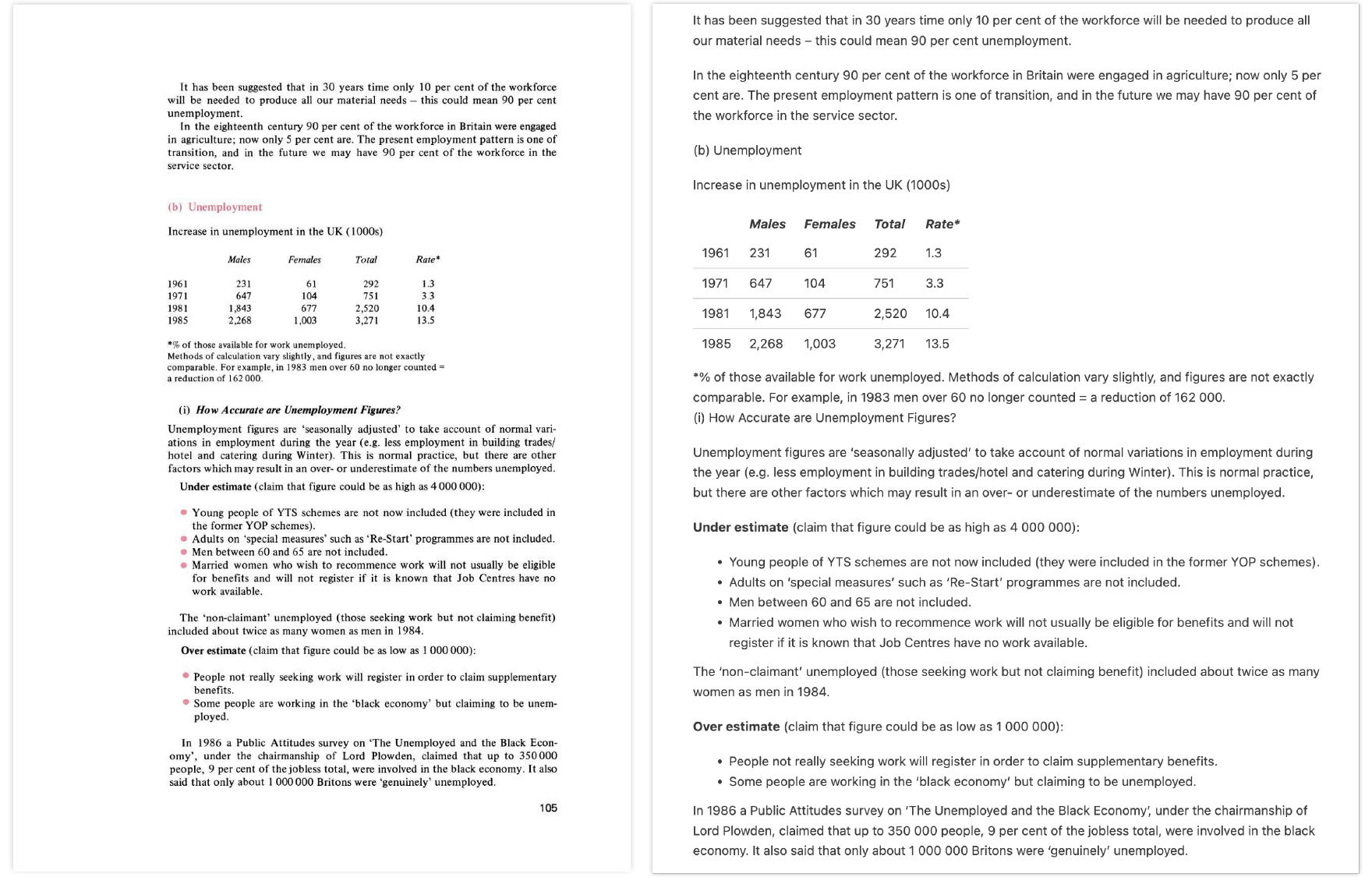}
\caption{The left image displays a document from DocMatix, while the right image presents the corresponding text, rendered as an image and automatically generated by our model}
\label{fig:af9}
\end{figure}

\begin{figure}[!htbp]
\centering
\includegraphics[width=\linewidth,scale=1.00]{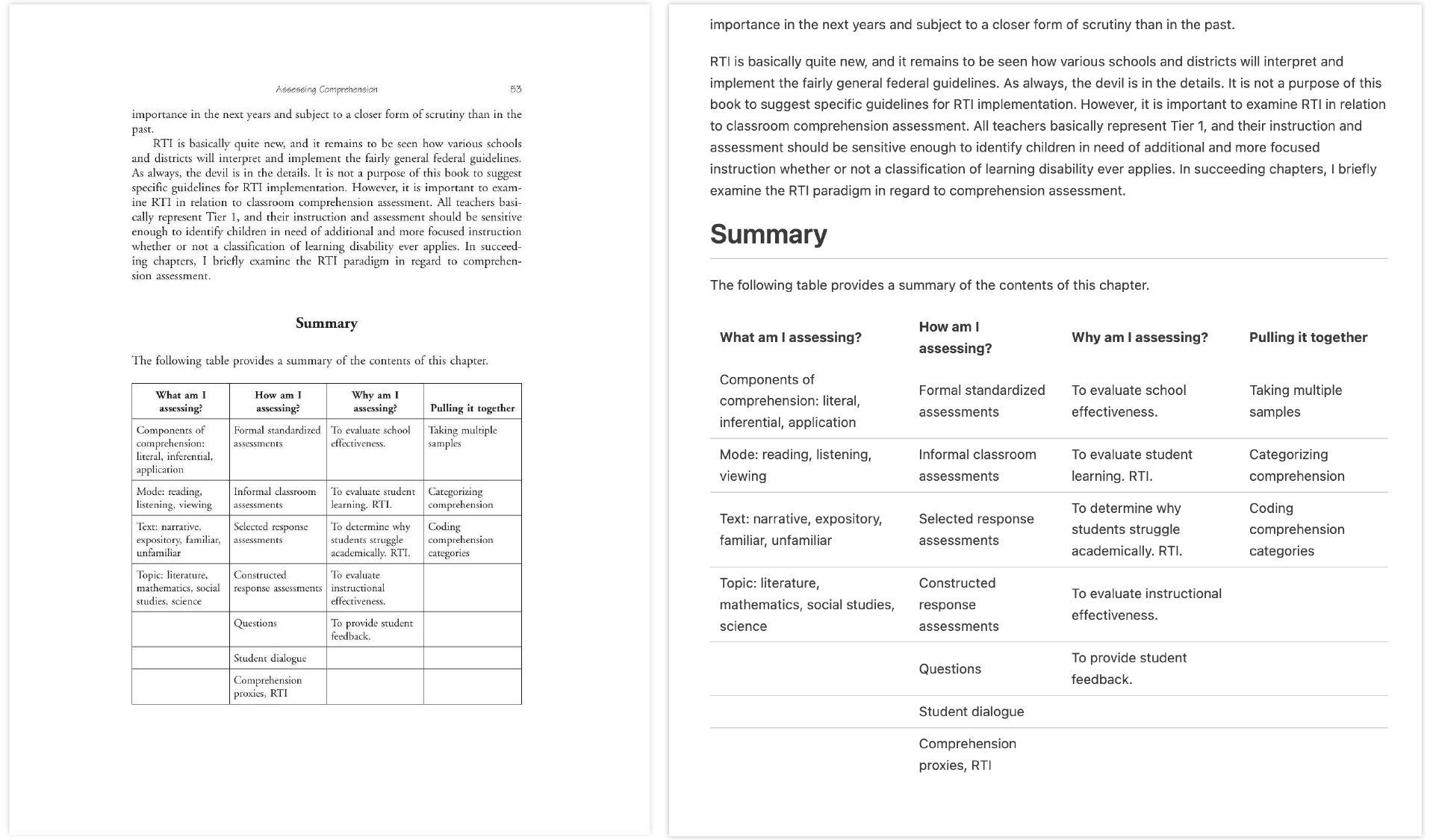}
\caption{The left image displays a document from DocMatix, while the right image presents the corresponding text, rendered as an image and automatically generated by our model}
\label{fig:af10}
\end{figure}

%% file: acl_latex.bbl
\begin{thebibliography}{43}
\providecommand{\natexlab}[1]{#1}

\bibitem[{Abdin et~al.(2024)Abdin, Aneja, Behl, Bubeck, Eldan, Gunasekar, Harrison, Hewett, Javaheripi, Kauffmann et~al.}]{abdin2024phi}
Marah Abdin, Jyoti Aneja, Harkirat Behl, S{\'e}bastien Bubeck, Ronen Eldan, Suriya Gunasekar, Michael Harrison, Russell~J Hewett, Mojan Javaheripi, Piero Kauffmann, and 1 others. 2024.
\newblock Phi-4 technical report.
\newblock \emph{arXiv preprint arXiv:2412.08905}.

\bibitem[{Bai et~al.(2025)Bai, Chen, Liu, Wang, Ge, Song, Dang, Wang, Wang, Tang et~al.}]{bai2025qwen2}
Shuai Bai, Keqin Chen, Xuejing Liu, Jialin Wang, Wenbin Ge, Sibo Song, Kai Dang, Peng Wang, Shijie Wang, Jun Tang, and 1 others. 2025.
\newblock Qwen2. 5-vl technical report.
\newblock \emph{arXiv preprint arXiv:2502.13923}.

\bibitem[{Blecher et~al.(2023)Blecher, Cucurull, Scialom, and Stojnic}]{blecher2023nougat}
Lukas Blecher, Guillem Cucurull, Thomas Scialom, and Robert Stojnic. 2023.
\newblock Nougat: Neural optical understanding for academic documents.
\newblock \emph{arXiv preprint arXiv:2308.13418}.

\bibitem[{Chen et~al.(2024)Chen, Wang, Cao, Liu, Gao, Cui, Zhu, Ye, Tian, Liu et~al.}]{chen2024expanding}
Zhe Chen, Weiyun Wang, Yue Cao, Yangzhou Liu, Zhangwei Gao, Erfei Cui, Jinguo Zhu, Shenglong Ye, Hao Tian, Zhaoyang Liu, and 1 others. 2024.
\newblock Expanding performance boundaries of open-source multimodal models with model, data, and test-time scaling.
\newblock \emph{arXiv preprint arXiv:2412.05271}.

\bibitem[{Cho et~al.(2025)Cho, Madotto, Mavroudi, Afouras, Nagarajan, Maaz, Song, Ma, Hu, Jain et~al.}]{cho2025perceptionlm}
Jang~Hyun Cho, Andrea Madotto, Effrosyni Mavroudi, Triantafyllos Afouras, Tushar Nagarajan, Muhammad Maaz, Yale Song, Tengyu Ma, Shuming Hu, Suyog Jain, and 1 others. 2025.
\newblock Perceptionlm: Open-access data and models for detailed visual understanding.
\newblock \emph{arXiv preprint arXiv:2504.13180}.

\bibitem[{Dehghani et~al.(2023)Dehghani, Mustafa, Djolonga, Heek, Minderer, Caron, Steiner, Puigcerver, Geirhos, Alabdulmohsin et~al.}]{dehghani2023patch}
Mostafa Dehghani, Basil Mustafa, Josip Djolonga, Jonathan Heek, Matthias Minderer, Mathilde Caron, Andreas Steiner, Joan Puigcerver, Robert Geirhos, Ibrahim~M Alabdulmohsin, and 1 others. 2023.
\newblock Patch n’pack: Navit, a vision transformer for any aspect ratio and resolution.
\newblock \emph{Advances in Neural Information Processing Systems}, 36:2252--2274.

\bibitem[{Du et~al.(2020)Du, Li, Guo, Yin, Liu, Zhou, Bai, Yu, Yang, Dang et~al.}]{du2020pp}
Yuning Du, Chenxia Li, Ruoyu Guo, Xiaoting Yin, Weiwei Liu, Jun Zhou, Yifan Bai, Zilin Yu, Yehua Yang, Qingqing Dang, and 1 others. 2020.
\newblock Pp-ocr: A practical ultra lightweight ocr system.
\newblock \emph{arXiv preprint arXiv:2009.09941}.

\bibitem[{Grattafiori et~al.(2024)Grattafiori, Dubey, Jauhri, Pandey, Kadian, Al-Dahle, Letman, Mathur, Schelten, Vaughan et~al.}]{grattafiori2024llama}
Aaron Grattafiori, Abhimanyu Dubey, Abhinav Jauhri, Abhinav Pandey, Abhishek Kadian, Ahmad Al-Dahle, Aiesha Letman, Akhil Mathur, Alan Schelten, Alex Vaughan, and 1 others. 2024.
\newblock The llama 3 herd of models.
\newblock \emph{arXiv preprint arXiv:2407.21783}.

\bibitem[{Guo et~al.(2025)Guo, Yang, Zhang, Song, Zhang, Xu, Zhu, Ma, Wang, Bi et~al.}]{guo2025deepseekr1}
Daya Guo, Dejian Yang, Haowei Zhang, Junxiao Song, Ruoyu Zhang, Runxin Xu, Qihao Zhu, Shirong Ma, Peiyi Wang, Xiao Bi, and 1 others. 2025.
\newblock Deepseek-r1: Incentivizing reasoning capability in llms via reinforcement learning.
\newblock \emph{arXiv preprint arXiv:2501.12948}.

\bibitem[{Hu et~al.(2024{\natexlab{a}})Hu, Xu, Ye, Yan, Zhang, Zhang, Zhang, Jin, Huang, and Zhou}]{hu-etal-2024-mplug}
Anwen Hu, Haiyang Xu, Jiabo Ye, Ming Yan, Liang Zhang, Bo~Zhang, Ji~Zhang, Qin Jin, Fei Huang, and Jingren Zhou. 2024{\natexlab{a}}.
\newblock \href {https://doi.org/10.18653/v1/2024.findings-emnlp.175} {m{PLUG}-{D}oc{O}wl 1.5: Unified structure learning for {OCR}-free document understanding}.
\newblock In \emph{Findings of the Association for Computational Linguistics: EMNLP 2024}, pages 3096--3120, Miami, Florida, USA. Association for Computational Linguistics.

\bibitem[{Hu et~al.(2024{\natexlab{b}})Hu, Xu, Zhang, Ye, Yan, Zhang, Jin, Huang, and Zhou}]{hu2024mplug2.0}
Anwen Hu, Haiyang Xu, Liang Zhang, Jiabo Ye, Ming Yan, Ji~Zhang, Qin Jin, Fei Huang, and Jingren Zhou. 2024{\natexlab{b}}.
\newblock mplug-docowl2: High-resolution compressing for ocr-free multi-page document understanding.
\newblock \emph{arXiv preprint arXiv:2409.03420}.

\bibitem[{Hurst et~al.(2024)Hurst, Lerer, Goucher, Perelman, Ramesh, Clark, Ostrow, Welihinda, Hayes, Radford et~al.}]{hurst2024gpt}
Aaron Hurst, Adam Lerer, Adam~P Goucher, Adam Perelman, Aditya Ramesh, Aidan Clark, AJ~Ostrow, Akila Welihinda, Alan Hayes, Alec Radford, and 1 others. 2024.
\newblock Gpt-4o system card.
\newblock \emph{arXiv preprint arXiv:2410.21276}.

\bibitem[{KaTex()}]{KaTex}
KaTex.

\bibitem[{Laurençon et~al.(2024)Laurençon, Marafioti, Sanh, and Tronchon}]{laurençon2024building}
Hugo Laurençon, Andrés Marafioti, Victor Sanh, and Léo Tronchon. 2024.
\newblock \href {https://arxiv.org/abs/2408.12637} {Building and better understanding vision-language models: insights and future directions.}
\newblock \emph{Preprint}, arXiv:2408.12637.

\bibitem[{Lee et~al.(2023)Lee, Joshi, Turc, Hu, Liu, Eisenschlos, Khandelwal, Shaw, Chang, and Toutanova}]{lee2023pix2struct}
Kenton Lee, Mandar Joshi, Iulia~Raluca Turc, Hexiang Hu, Fangyu Liu, Julian~Martin Eisenschlos, Urvashi Khandelwal, Peter Shaw, Ming-Wei Chang, and Kristina Toutanova. 2023.
\newblock Pix2struct: Screenshot parsing as pretraining for visual language understanding.
\newblock In \emph{International Conference on Machine Learning}, pages 18893--18912. PMLR.

\bibitem[{Li et~al.(2024)Li, Zhang, Guo, Zhang, Li, Zhang, Zhang, Zhang, Li, Liu et~al.}]{li2024llava}
Bo~Li, Yuanhan Zhang, Dong Guo, Renrui Zhang, Feng Li, Hao Zhang, Kaichen Zhang, Peiyuan Zhang, Yanwei Li, Ziwei Liu, and 1 others. 2024.
\newblock Llava-onevision: Easy visual task transfer.
\newblock \emph{arXiv preprint arXiv:2408.03326}.

\bibitem[{Li et~al.(2022)Li, Guo, Zhou, An, Du, Zhu, Liu, Hu, and Yu}]{li2022slanet}
Chenxia Li, Ruoyu Guo, Jun Zhou, Mengtao An, Yuning Du, Lingfeng Zhu, Yi~Liu, Xiaoguang Hu, and Dianhai Yu. 2022.
\newblock Pp-structurev2: A stronger document analysis system.
\newblock \emph{arXiv preprint arXiv:2210.05391}.

\bibitem[{Li et~al.(2023)Li, Li, Savarese, and Hoi}]{li2023blip}
Junnan Li, Dongxu Li, Silvio Savarese, and Steven Hoi. 2023.
\newblock Blip-2: Bootstrapping language-image pre-training with frozen image encoders and large language models.
\newblock In \emph{International conference on machine learning}, pages 19730--19742. PMLR.

\bibitem[{Liu et~al.(2024{\natexlab{a}})Liu, Feng, Xue, Wang, Wu, Lu, Zhao, Deng, Zhang, Ruan et~al.}]{liu2024deepseekv3}
Aixin Liu, Bei Feng, Bing Xue, Bingxuan Wang, Bochao Wu, Chengda Lu, Chenggang Zhao, Chengqi Deng, Chenyu Zhang, Chong Ruan, and 1 others. 2024{\natexlab{a}}.
\newblock Deepseek-v3 technical report.
\newblock \emph{arXiv preprint arXiv:2412.19437}.

\bibitem[{Liu et~al.(2024{\natexlab{b}})Liu, Wei, Chen, Kong, Ge, Zhu, Zhao, Sun, Han, and Zhang}]{liu2024focus}
Chenglong Liu, Haoran Wei, Jinyue Chen, Lingyu Kong, Zheng Ge, Zining Zhu, Liang Zhao, Jianjian Sun, Chunrui Han, and Xiangyu Zhang. 2024{\natexlab{b}}.
\newblock Focus anywhere for fine-grained multi-page document understanding.
\newblock \emph{arXiv preprint arXiv:2405.14295}.

\bibitem[{Liu et~al.(2024{\natexlab{c}})Liu, Li, Li, and Lee}]{liu2024improved}
Haotian Liu, Chunyuan Li, Yuheng Li, and Yong~Jae Lee. 2024{\natexlab{c}}.
\newblock Improved baselines with visual instruction tuning.
\newblock In \emph{Proceedings of the IEEE/CVF Conference on Computer Vision and Pattern Recognition}, pages 26296--26306.

\bibitem[{Liu et~al.(2023)Liu, Li, Wu, and Lee}]{liu2023visual}
Haotian Liu, Chunyuan Li, Qingyang Wu, and Yong~Jae Lee. 2023.
\newblock Visual instruction tuning.
\newblock \emph{Advances in neural information processing systems}, 36:34892--34916.

\bibitem[{Liu et~al.(2024{\natexlab{d}})Liu, Duan, Zhang, Li, Zhang, Zhao, Yuan, Wang, He, Liu et~al.}]{liu2024mmbench}
Yuan Liu, Haodong Duan, Yuanhan Zhang, Bo~Li, Songyang Zhang, Wangbo Zhao, Yike Yuan, Jiaqi Wang, Conghui He, Ziwei Liu, and 1 others. 2024{\natexlab{d}}.
\newblock Mmbench: Is your multi-modal model an all-around player?
\newblock In \emph{European conference on computer vision}, pages 216--233. Springer.

\bibitem[{Liu et~al.(2024{\natexlab{e}})Liu, Tian, Zhou, Gao, Yu, Yu, and Zhou}]{liu2024points1}
Yuan Liu, Le~Tian, Xiao Zhou, Xinyu Gao, Kavio Yu, Yang Yu, and Jie Zhou. 2024{\natexlab{e}}.
\newblock Points1. 5: Building a vision-language model towards real world applications.
\newblock \emph{arXiv preprint arXiv:2412.08443}.

\bibitem[{Liu et~al.(2024{\natexlab{f}})Liu, Tian, Zhou, and Zhou}]{liu2024rethinking}
Yuan Liu, Le~Tian, Xiao Zhou, and Jie Zhou. 2024{\natexlab{f}}.
\newblock Rethinking overlooked aspects in vision-language models.
\newblock \emph{arXiv preprint arXiv:2405.11850}.

\bibitem[{Liu et~al.(2024{\natexlab{g}})Liu, Zhao, Zhuang, Tian, Zhou, and Zhou}]{liu2024points}
Yuan Liu, Zhongyin Zhao, Ziyuan Zhuang, Le~Tian, Xiao Zhou, and Jie Zhou. 2024{\natexlab{g}}.
\newblock Points: Improving your vision-language model with affordable strategies.
\newblock \emph{arXiv preprint arXiv:2409.04828}.

\bibitem[{Lv et~al.(2023)Lv, Huang, Chen, Zhao, Jia, Cui, Ma, Chang, Huang, Wang et~al.}]{lv2023kosmos}
Tengchao Lv, Yupan Huang, Jingye Chen, Yuzhong Zhao, Yilin Jia, Lei Cui, Shuming Ma, Yaoyao Chang, Shaohan Huang, Wenhui Wang, and 1 others. 2023.
\newblock Kosmos-2.5: A multimodal literate model.
\newblock \emph{arXiv preprint arXiv:2309.11419}.

\bibitem[{marker(2024)}]{marker}
marker. 2024.
\newblock \url{https://github.com/VikParuchuri/marker}.

\bibitem[{mathpix(2024)}]{mathpix}
mathpix. 2024.
\newblock \url{https://github.com/Mathpix/mathpix-markdown-it}.

\bibitem[{Nassar et~al.(2025)Nassar, Marafioti, Omenetti, Lysak, Livathinos, Auer, Morin, de~Lima, Kim, Gurbuz et~al.}]{nassar2025smoldocling}
Ahmed Nassar, Andres Marafioti, Matteo Omenetti, Maksym Lysak, Nikolaos Livathinos, Christoph Auer, Lucas Morin, Rafael~Teixeira de~Lima, Yusik Kim, A~Said Gurbuz, and 1 others. 2025.
\newblock Smoldocling: An ultra-compact vision-language model for end-to-end multi-modal document conversion.
\newblock \emph{arXiv preprint arXiv:2503.11576}.

\bibitem[{Ouyang et~al.(2024)Ouyang, Qu, Zhou, Zhu, Zhang, Lin, Wang, Zhao, Jiang, Zhao, Shi, Wu, Chu, Liu, Li, Xu, Zhang, Shi, Tu, and He}]{ouyang2024omnidocbenchbenchmarkingdiversepdf}
Linke Ouyang, Yuan Qu, Hongbin Zhou, Jiawei Zhu, Rui Zhang, Qunshu Lin, Bin Wang, Zhiyuan Zhao, Man Jiang, Xiaomeng Zhao, Jin Shi, Fan Wu, Pei Chu, Minghao Liu, Zhenxiang Li, Chao Xu, Bo~Zhang, Botian Shi, Zhongying Tu, and Conghui He. 2024.
\newblock \href {https://arxiv.org/abs/2412.07626} {Omnidocbench: Benchmarking diverse pdf document parsing with comprehensive annotations}.
\newblock \emph{Preprint}, arXiv:2412.07626.

\bibitem[{Poznanski et~al.(2025)Poznanski, Borchardt, Dunkelberger, Huff, Lin, Rangapur, Wilhelm, Lo, and Soldaini}]{poznanski2025olmocr}
Jake Poznanski, Jon Borchardt, Jason Dunkelberger, Regan Huff, Daniel Lin, Aman Rangapur, Christopher Wilhelm, Kyle Lo, and Luca Soldaini. 2025.
\newblock olmocr: Unlocking trillions of tokens in pdfs with vision language models.
\newblock \emph{arXiv preprint arXiv:2502.18443}.

\bibitem[{Team(2024)}]{Docling}
Deep~Search Team. 2024.
\newblock \href {https://arxiv.org/abs/2408.09869} {Docling technical report}.
\newblock \emph{10.48550/arXiv.2408.09869}.

\bibitem[{Team et~al.(2025)Team, Du, Yin, Xing, Qu, Wang, Chen, Zhang, Du, Wei et~al.}]{team2025kimivl}
Kimi Team, Angang Du, Bohong Yin, Bowei Xing, Bowen Qu, Bowen Wang, Cheng Chen, Chenlin Zhang, Chenzhuang Du, Chu Wei, and 1 others. 2025.
\newblock Kimi-vl technical report.
\newblock \emph{arXiv preprint arXiv:2504.07491}.

\bibitem[{Wang et~al.(2024{\natexlab{a}})Wang, Xu, Zhao, Ouyang, Wu, Zhao, Xu, Liu, Qu, Shang et~al.}]{wang2024mineru}
Bin Wang, Chao Xu, Xiaomeng Zhao, Linke Ouyang, Fan Wu, Zhiyuan Zhao, Rui Xu, Kaiwen Liu, Yuan Qu, Fukai Shang, and 1 others. 2024{\natexlab{a}}.
\newblock Mineru: An open-source solution for precise document content extraction.
\newblock \emph{arXiv preprint arXiv:2409.18839}.

\bibitem[{Wang et~al.(2024{\natexlab{b}})Wang, Bai, Tan, Wang, Fan, Bai, Chen, Liu, Wang, Ge et~al.}]{wang2024qwen2}
Peng Wang, Shuai Bai, Sinan Tan, Shijie Wang, Zhihao Fan, Jinze Bai, Keqin Chen, Xuejing Liu, Jialin Wang, Wenbin Ge, and 1 others. 2024{\natexlab{b}}.
\newblock Qwen2-vl: Enhancing vision-language model's perception of the world at any resolution.
\newblock \emph{arXiv preprint arXiv:2409.12191}.

\bibitem[{Wei et~al.(2024)Wei, Liu, Chen, Wang, Kong, Xu, Ge, Zhao, Sun, Peng et~al.}]{wei2024general}
Haoran Wei, Chenglong Liu, Jinyue Chen, Jia Wang, Lingyu Kong, Yanming Xu, Zheng Ge, Liang Zhao, Jianjian Sun, Yuang Peng, and 1 others. 2024.
\newblock General ocr theory: Towards ocr-2.0 via a unified end-to-end model.
\newblock \emph{arXiv preprint arXiv:2409.01704}.

\bibitem[{Xia et~al.(2024)Xia, Mao, Yan, Zhou, Zhang, Peng, Pi, Fu, Wu, Ye et~al.}]{xia2024docgenome}
Renqiu Xia, Song Mao, Xiangchao Yan, Hongbin Zhou, Bo~Zhang, Haoyang Peng, Jiahao Pi, Daocheng Fu, Wenjie Wu, Hancheng Ye, and 1 others. 2024.
\newblock Docgenome: An open large-scale scientific document benchmark for training and testing multi-modal large language models.
\newblock \emph{arXiv preprint arXiv:2406.11633}.

\bibitem[{Yang et~al.(2024{\natexlab{a}})Yang, Yang, Zhang, Hui, Zheng, Yu, Li, Liu, Huang, Wei et~al.}]{yang2024qwen2}
An~Yang, Baosong Yang, Beichen Zhang, Binyuan Hui, Bo~Zheng, Bowen Yu, Chengyuan Li, Dayiheng Liu, Fei Huang, Haoran Wei, and 1 others. 2024{\natexlab{a}}.
\newblock Qwen2. 5 technical report.
\newblock \emph{arXiv preprint arXiv:2412.15115}.

\bibitem[{Yang et~al.(2024{\natexlab{b}})Yang, Tang, Li, Wang, Wan, Zhong, Liu, Yang, Wang, Liu et~al.}]{yang2024cc}
Zhibo Yang, Jun Tang, Zhaohai Li, Pengfei Wang, Jianqiang Wan, Humen Zhong, Xuejing Liu, Mingkun Yang, Peng Wang, Yuliang Liu, and 1 others. 2024{\natexlab{b}}.
\newblock Cc-ocr: A comprehensive and challenging ocr benchmark for evaluating large multimodal models in literacy.
\newblock \emph{arXiv preprint arXiv:2412.02210}.

\bibitem[{Zhang et~al.(2023)Zhang, Dong, Wang, Cao, Xu, Ouyang, Zhao, Duan, Zhang, Ding et~al.}]{zhang2023internlm}
Pan Zhang, Xiaoyi Dong, Bin Wang, Yuhang Cao, Chao Xu, Linke Ouyang, Zhiyuan Zhao, Haodong Duan, Songyang Zhang, Shuangrui Ding, and 1 others. 2023.
\newblock Internlm-xcomposer: A vision-language large model for advanced text-image comprehension and composition.
\newblock \emph{arXiv preprint arXiv:2309.15112}.

\bibitem[{Zhang et~al.(2024)Zhang, Dong, Zang, Cao, Qian, Chen, Guo, Duan, Wang, Ouyang et~al.}]{zhang2024internlm}
Pan Zhang, Xiaoyi Dong, Yuhang Zang, Yuhang Cao, Rui Qian, Lin Chen, Qipeng Guo, Haodong Duan, Bin Wang, Linke Ouyang, and 1 others. 2024.
\newblock Internlm-xcomposer-2.5: A versatile large vision language model supporting long-contextual input and output.
\newblock \emph{arXiv preprint arXiv:2407.03320}.

\bibitem[{Zhong et~al.(2020)Zhong, ShafieiBavani, and Jimeno~Yepes}]{zhong2020image}
Xu~Zhong, Elaheh ShafieiBavani, and Antonio Jimeno~Yepes. 2020.
\newblock Image-based table recognition: data, model, and evaluation.
\newblock In \emph{European conference on computer vision}, pages 564--580. Springer.

\end{thebibliography}
